\documentclass[11pt,twoside,leqno]{article}
\usepackage{jair,theapa}
\usepackage{times,helvet,courier}

\usepackage{alltt}
\usepackage{graphicx}
\usepackage{psfrag}
\usepackage{amsmath,amssymb}
\newtheorem{def1}{Definition}
\newtheorem{obs1}[def1]{Observation}

\newenvironment{definition}[2]
  {\begin{def1}{\label{def:#1}}\definedTerm{#2}\\ }
  {\end{def1}}
\newenvironment{observation}[2]
  {\begin{obs1}{\label{observation:#1}}\definedTerm{#2}\\ }
  {\end{obs1}}

\newenvironment{algorithm}
  {\begin{upshape}\setlength{\parindent}{0cm}}
  {\end{upshape}}

\newcommand{\includepsfigure}[3][tbp]{
  \begin{figure}[#1]
    \begin{center}
      \includegraphics{figures/#2}
    \end{center}
    \caption{#3}
    \label{figure:#2}
  \end{figure}
}

\newcommand{\includegraph}[3][tbp]{
  \begin{figure}[#1]
    \begin{center}
      \includegraphics{figures/#2}
    \end{center}
    \caption{#3}
    \label{figure:#2}
  \end{figure}
}

\newcommand{\ie}{i.\,e.\@}

\newcommand{\pre}                   {\textit{pre}}
\newcommand{\eff}                   {\textit{eff}}
\newcommand{\cond}                  {\textit{cond}}

\newcommand{\decisionProblem}[1] {\textsc{#1}}
\newcommand{\planningDomain}[1]  {\textsc{#1}}

\newcommand{\definedTerm}[1]     {\textbf{#1}}

\newcommand{\sasplus}{SAS\ensuremath{{}^+}}

\newcommand{\NP}                 {\textbf{\textup{NP}}}
\newcommand{\PSPACE}             {\textbf{\textup{PSPACE}}}

\newcommand{\mptplanex}          {\decisionProblem{MPT-PlanEx}}
\newcommand{\mptplan}            {\decisionProblem{MPT-Planning}}

\newcommand{\airport}            {\planningDomain{Airport}}
\newcommand{\assembly}           {\planningDomain{Assembly}}
\newcommand{\blocksworld}        {\planningDomain{Blocksworld}}
\newcommand{\depot}              {\planningDomain{Depot}}
\newcommand{\driverlog}          {\planningDomain{Driverlog}}
\newcommand{\freecell}           {\planningDomain{Freecell}}
\newcommand{\grid}               {\planningDomain{Grid}}
\newcommand{\gripper}            {\planningDomain{Gripper}}
\newcommand{\logistics}          {\planningDomain{Logistics}}
\newcommand{\miconic}            {\planningDomain{Miconic}}
\newcommand{\miconicstrips}      {\planningDomain{Miconic-STRIPS}}
\newcommand{\miconicsimple}      {\planningDomain{Miconic-SimpleADL}}
\newcommand{\miconicfull}        {\planningDomain{Miconic-FullADL}}
\newcommand{\movie}              {\planningDomain{Movie}}
\newcommand{\mystery}            {\planningDomain{Mystery}}
\newcommand{\mysteryprime}       {\planningDomain{MPrime}}
\newcommand{\opticaltelegraph}   {\planningDomain{Promela-OpticalTelegraph}}
\newcommand{\philosophers}       {\planningDomain{Promela-Philosophers}}
\newcommand{\pipesworld}         {\planningDomain{Pipesworld}}
\newcommand{\pipesworldnotankage}{\planningDomain{Pipesworld-NoTankage}}
\newcommand{\pipesworldtankage}  {\planningDomain{Pipesworld-Tankage}}
\newcommand{\promela}            {\planningDomain{Promela}}
\newcommand{\psr}                {\planningDomain{PSR}}
\newcommand{\psrsmall}           {\planningDomain{PSR-Small}}
\newcommand{\psrmiddle}          {\planningDomain{PSR-Middle}}
\newcommand{\psrlarge}           {\planningDomain{PSR-Large}}
\newcommand{\rovers}             {\planningDomain{Rovers}}
\newcommand{\satellite}          {\planningDomain{Satellite}}
\newcommand{\schedule}           {\planningDomain{Schedule}}
\newcommand{\zenotravel}         {\planningDomain{Zenotravel}}

\ShortHeadings{The Fast Downward Planning System}{Helmert}

\jairheading{26}{2006}{191--246}{01/05}{07/06}
\firstpageno{191}

\begin{document}
\title{The Fast Downward Planning System}
\author{\name Malte Helmert
        \email helmert@informatik.uni-freiburg.de \\
        \addr Institut f\"ur Informatik \\
          Albert-Ludwigs-Universit\"at Freiburg \\
          Georges-K\"ohler-Allee, Geb\"aude 052 \\
          79110 Freiburg, Germany}

\maketitle

\begin{abstract}
  Fast Downward is a classical planning system based on heuristic
  search. It can deal with general deterministic planning problems
  encoded in the propositional fragment of PDDL2.2, including advanced
  features like ADL conditions and effects and derived predicates
  (axioms). Like other well-known planners such as HSP and
  FF, Fast Downward is a progression planner, searching the
  space of world states of a planning task in the forward direction.
  However, unlike other PDDL planning systems, Fast Downward does not
  use the propositional PDDL representation of a planning task directly.
  Instead, the input is first translated into an alternative
  representation called \emph{multi-valued planning tasks}, which
  makes many of the implicit constraints of a propositional planning
  task explicit. Exploiting this alternative representation, Fast
  Downward uses hierarchical decompositions of planning tasks for
  computing its heuristic function, called the \emph{causal graph
  heuristic}, which is very different from traditional HSP-like
  heuristics based on ignoring negative interactions of operators.

  In this article, we give a full account of Fast Downward's approach
  to solving multi-valued planning tasks. We extend our earlier
  discussion of the causal graph heuristic to
  tasks involving axioms and conditional effects and present some
  novel techniques for search control that are used within Fast
  Downward's best-first search algorithm: \emph{preferred operators}
  transfer the idea of helpful actions from local search to
  global best-first search, \emph{deferred evaluation} of heuristic
  functions mitigates the negative effect of large branching factors
  on search performance, and \emph{multi-heuristic best-first search}
  combines several heuristic evaluation functions within a single
  search algorithm in an orthogonal way. We also describe efficient
  data structures for fast state expansion (\emph{successor
  generators} and \emph{axiom evaluators}) and present a new
  non-heuristic search algorithm called \emph{focused
  iterative-broadening search}, which utilizes the information encoded
  in causal graphs in a novel way.

  Fast Downward has proven remarkably successful: It won the
  ``classical'' ({\ie}, propositional, non-optimising) track of the 4th
  International Planning Competition at ICAPS 2004, following in the
  footsteps of planners such as FF and LPG. Our
  experiments show that it also performs very well on the benchmarks
  of the earlier planning competitions and provide some insights about
  the usefulness of the new search enhancements.
\end{abstract}

\section{Introduction}
\label{section:introduction}

Consider a typical transportation planning task: The postal service must
deliver a number of parcels to their respective destinations using its
vehicle fleet of cars and trucks. Let us assume that a car serves all
the locations of one city, and that different cities are connected via
highways that are served by trucks. For the sake of simplicity, let us
further assume that travelling on each segment of road or highway incurs
the same cost. This is not a highly realistic assumption, but for the
purposes of exposition it will do. There can be any number of parcels,
posted at arbitrary locations and with arbitrary destinations. Moreover,
cities can be of varying size, there can be one or several cars within
each city, and there can be one or several trucks connecting the
cities. Cars will never leave a city.  Fig.~\ref{figure:initial-example}
shows an example task of this kind with two cities, three cars and a
single truck. There are two parcels to be delivered, one of which
($p_1$) must be moved between the two cities, while the other ($p_2$)
can stay within its initial city.

\includepsfigure{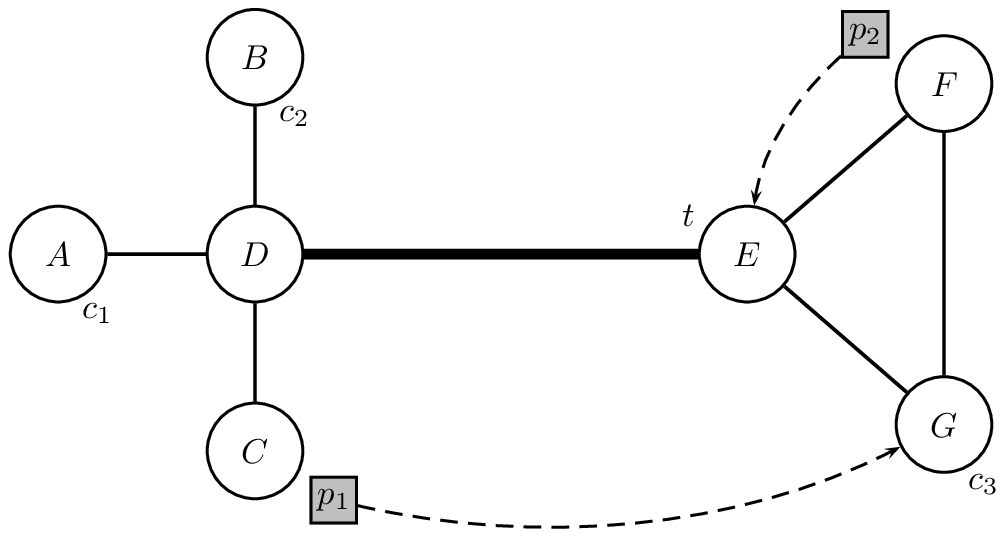}{A transportation planning task. Deliver
parcel $p_1$ from $C$ to $G$ and parcel $p_2$ from $F$ to $E$, using the
cars $c_1$, $c_2$, $c_3$ and truck $t$. The cars may only use inner-city
roads (thin edges), the truck may only use the highway (thick edge).}

The astute reader familiar with the planning literature will have
noticed by now that we are essentially describing the {\logistics}
domain, a standard benchmark for classical planning systems, extended to
roadmaps that are not complete graphs. (Part of) a propositional
STRIPS-like encoding of the task is shown in
Fig.~\ref{figure:initial-example-strips}.

  \begin{figure}[p]
    \begin{center}
      \input{figures/initial-example-strips}
    \end{center}
    \caption{Part of a typical
propositional encoding of the transportation planning task (no actual
PDDL syntax).}
    \label{figure:initial-example-strips}
  \end{figure}

How would human planners go about solving tasks of this kind? Very
likely, they would use a hierarchical approach: For $p_1$, it is clear
that the parcel needs to be moved between cities, which is only possible
by using the truck. Since in our example each city can access the
highway at only one location, we see that we must first load the parcel
into some car at its initial location, then drop it off at the first
city's highway access location, load it into the truck, drop it off at
the other city's highway access location, load it into the only car in
that city, and finally drop it off at its destination. We can commit to
this ``high-level'' plan for delivering $p_1$ without worrying about
``lower-level'' aspects such as path planning for the cars. It is
obvious to us that \emph{any} good solution will have this structure,
since the parcel can only change its location in a few clearly defined
ways (Fig.~\ref{figure:initial-example-parcel-dtg}). The same figure
shows that the only reasonable plans for getting $p_2$ to its
destination require loading it into the car in its initial city and
dropping it off at its target location. There is no point in ever
loading it into the truck or into any of the cars in the left city.

\includepsfigure{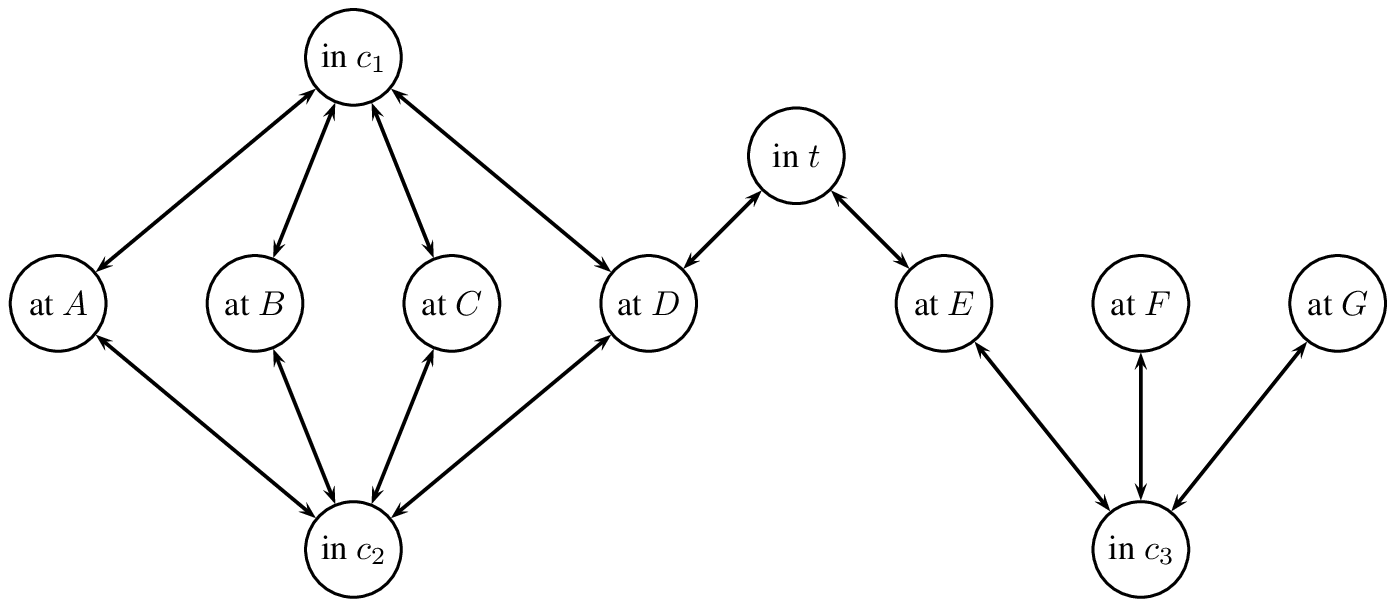}{Domain transition graph for
the parcels $p_1$ and $p_2$. Indicates how a parcel can change its
state. For example, the arcs between ``at $D$'' and ``in $t$''
correspond to the actions of loading/unloading the parcel at location
$D$ with the truck $t$.}

\includepsfigure{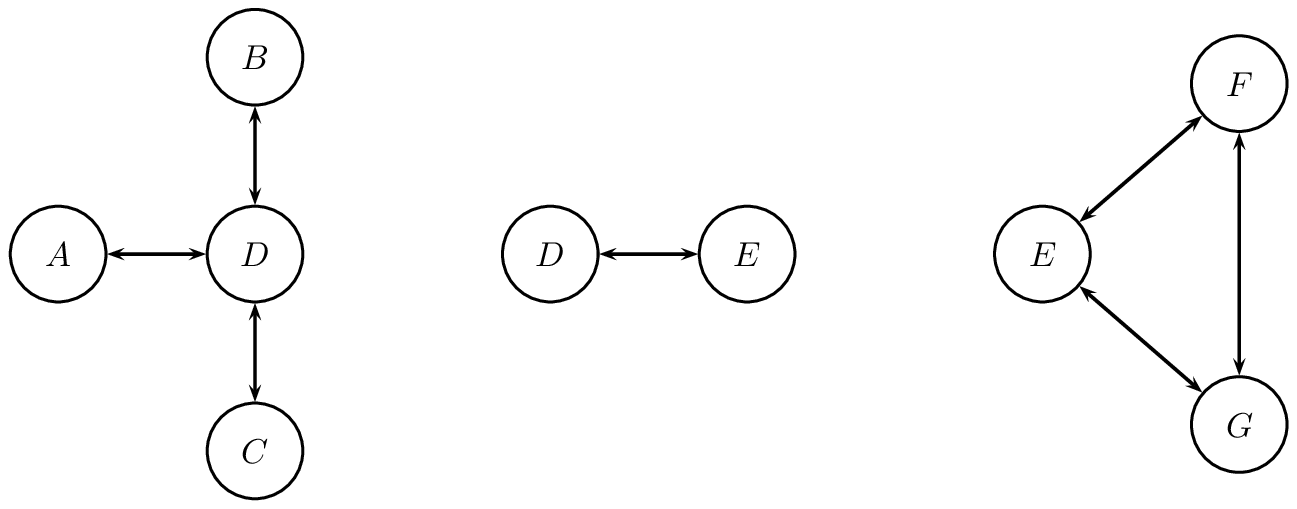}{Domain transition graphs
for the cars $c_1$ and $c_2$ (left), truck $t$ (centre), and car $c_3$
(right). Note how each graph corresponds to the part of the roadmap that can
be traversed by the respective vehicle.}

\includepsfigure{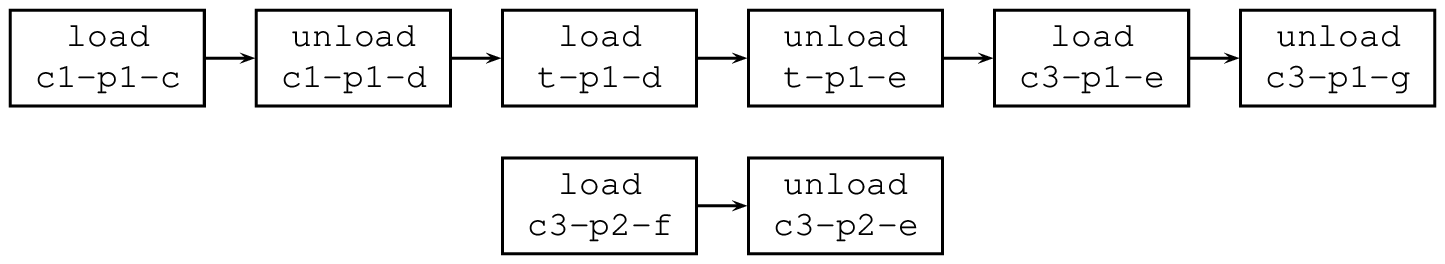}{High-level plan for
the transportation planning task.}

So say we have committed to the (partially ordered, as movements of the
two parcels can be interleaved) ``high-level plan'' shown in
Fig.~\ref{figure:initial-example-high-level-plan}. All we need to do to
complete the plan is choose a linearization of the high-level steps and
fill in movements of the vehicle fleet between them. We have thus
decomposed the planning task into a number of subproblems. The parcel
scheduling problem (where, and by which vehicles, a parcel should be
loaded and unloaded) is separated from the path planning problem for
each vehicle in the fleet (how to move it from point X to Y). Both of
these are graph search problems, and the corresponding graphs are shown
in Fig.~\ref{figure:initial-example-parcel-dtg} and
Fig.~\ref{figure:initial-example-other-dtgs}. Graphs of this kind will
be formally introduced as \emph{domain transition graphs} in Section
\ref{section:knowledge-compilation}.

Of course these graph search problems interact, but they only do so in
limited ways: State transitions for the parcels have associated
conditions regarding the vehicle fleet, which need to be considered in
addition to the actual path planning in
Fig.~\ref{figure:initial-example-parcel-dtg}. For example, a parcel can
only change state from ``at location $A$'' to ``inside car $c_1$'' if
the car $c_1$ is at location $A$. However, state transitions for the
vehicles have no associated conditions from other parts of the planning
task, and hence moving a vehicle from one location to another is indeed
as easy as finding a path in the associated domain transition graph. We
say that the parcels have \emph{causal dependencies} on the vehicles
because there are operators that change the state of the parcels and
have preconditions on the state of the vehicles. Indeed, these are the
only causal dependencies in this task, since parcels do not depend on
other parcels and vehicles do not depend on anything except themselves
(Fig.~\ref{figure:initial-example-causal-graph}). The set of causal
dependencies of a planning task is visualized in its
\emph{causal graph}.

\includepsfigure{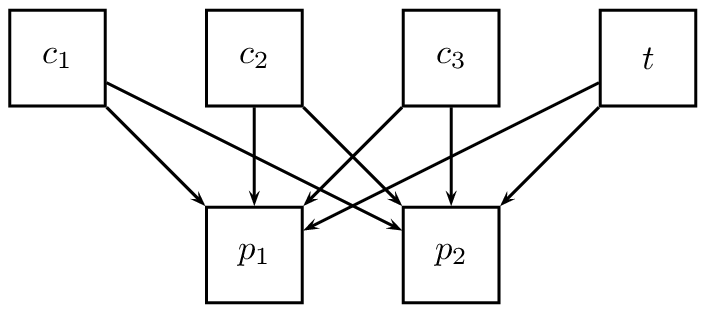}{Causal dependencies in
the transportation planning task.}

We argue that humans often solve planning tasks in the hierarchical
fashion outlined in the preceding paragraphs, and that algorithmic
approaches to action planning can usefully apply similar ideas.
Indeed, as we will show in the following section, we are not the first
to introduce domain transition graphs and causal graphs. However,
earlier work has almost exclusively focused on \emph{acyclic} causal
graphs, and for a good reason: If the causal graph of a planning task
exhibits a cycle, hierarchical decomposition is not possible, because
the subproblems that must be solved to achieve an operator
precondition are not necessarily smaller than the original task. As
far as we are aware, we were the first \cite{malte-icaps-cg} to
present a \emph{general} planning algorithm that focuses on exploiting
hierarchical information from causal graphs. However, our \emph{causal
graph heuristic} also requires acyclicity; in the general case, it
considers a relaxed planning problem in which some operator
preconditions are ignored to break causal cycles.

Knowing that cycles in causal graphs are undesirable, we take a closer
look at the transportation planning task. Let us recall our informal
definition of causal graphs: The causal graph of a planning task
contains a vertex for each state variable and arcs from variables that
occur in preconditions to variables that occur in effects of the same
operator. So far, we may have given the impression that the causal graph
of the example task has the well-behaved shape shown in
Fig.~\ref{figure:initial-example-causal-graph}. Unfortunately, having a
closer look at the STRIPS encoding in
Fig.~\ref{figure:initial-example-strips}, we see that this is not the
case: The correct causal graph, shown in
Fig.~\ref{figure:initial-example-strips-causal-graph}, looks very messy.
This discrepancy between the intuitive and actual graph is due to the
fact that in our informal account of ``human-style'' problem solving, we
made use of (non-binary) state variables like ``the location of car
$c_1$'' or ``the state of parcel $p_1$'', while STRIPS-level state
variables correspond to (binary) object-location propositions like
``parcel $p_1$ is at location $A$''. It would be much nicer if we were
given a multi-valued encoding of the planning task that explicitly
contains a variable for ``the location of car $c_1$'' and similar
properties. Indeed, the nice looking acyclic graph in
Fig.~\ref{figure:initial-example-causal-graph} is the causal graph of
the multi-valued encoding shown in
Fig.~\ref{figure:initial-example-sas}.

\includepsfigure{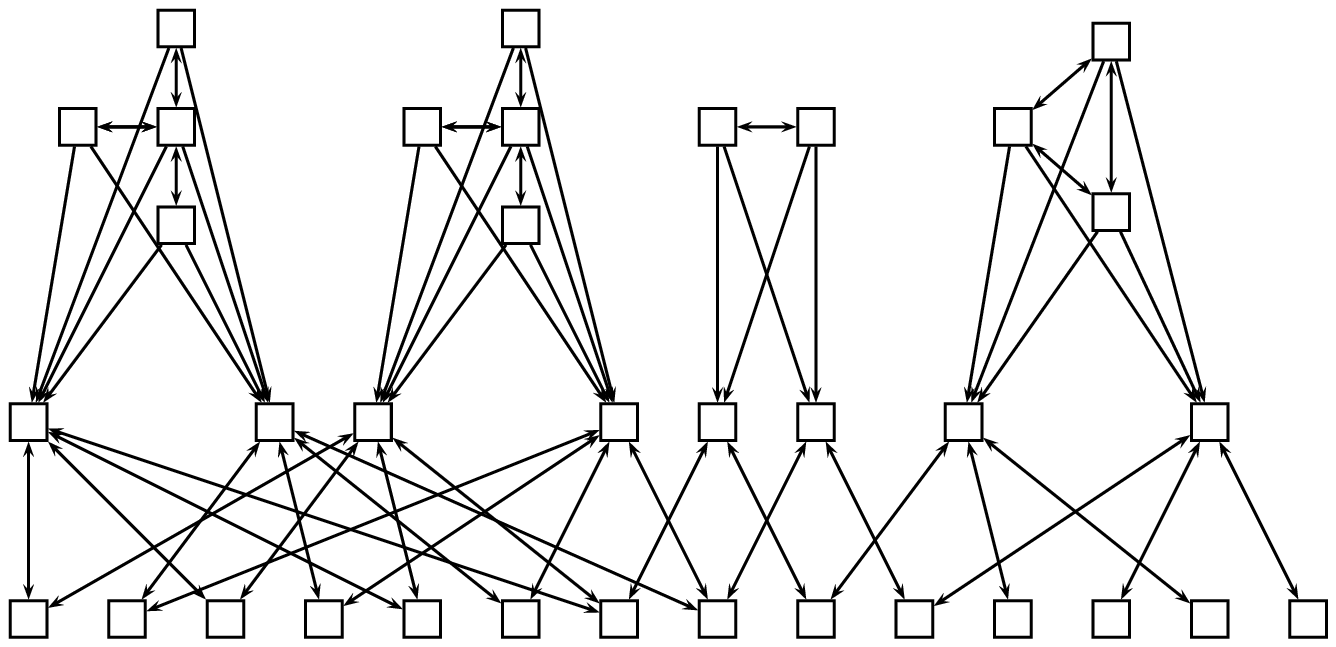}{Causal graph for
the STRIPS encoding of the transportation planning task.}

  \begin{figure}[p]
    \begin{center}
      \input{figures/initial-example-sas}
    \end{center}
    \caption{Part of an encoding of the
transportation planning task with multi-valued state variables.}
    \label{figure:initial-example-sas}
  \end{figure}

Having provided some intuition for its underlying concepts, let us now
state our design goal for the Fast Downward planning system: \emph{To
develop an algorithm that efficiently solves general propositional
planning tasks by exploiting the hierarchical structure inherent in
causal graphs.} We need to overcome three major obstacles in
this undertaking:
\begin{itemize}
\item First, propositionally encoded planning tasks usually have very
  unstructured causal graphs. However, the intuitive dependencies
  often become visible in encodings with multi-valued state variables.
  To exploit this fact in an automated PDDL planning system, we have
  devised an automatic algorithm for ``translating'' (or
  reformulating) propositional tasks to multi-valued ones. The
  translation algorithm can be considered independently from the rest
  of the planner; in fact, it is now also used as part of other
  planning systems \cite{menkes-ilp}. To keep the article focused, we
  do \emph{not} discuss the translation algorithm here, referring to
  our earlier work for some of its central ideas \cite{minimize}.
  Instead, we consider its output, a \emph{multi-valued planning
    task}, as a base formalism.
\item Second, no matter how clever the encoding is, most planning
  tasks are not completely hierarchical in nature. To deal with causal
  cycles, we consider relaxations where some causal dependencies are
  ignored and use solutions to the relaxed problem within a heuristic
  search algorithm.
\item Third, even for planning tasks that can be solved
  hierarchically, finding such a solution is difficult (indeed, still
  \PSPACE-complete). For this reason, our heuristic function only
  considers a fragment of a task at a time, namely subproblems induced
  by a single state variable and its predecessors in the causal graph.
  Even \emph{this} planning problem is still NP-complete, so that we
  are content with an incomplete solution algorithm within the
  heuristic solver. This solution algorithm has theoretical
  shortcomings but never failed us in practice.
\end{itemize}

Having introduced the rationale of our approach, we discuss related
work in the next section. This is followed by an overview of the
general architecture of the Fast Downward planning system in
Section~\ref{section:architecture}. The planning system consists of
three components: \emph{translation}, \emph{knowledge compilation},
and \emph{search}. The translation component converts PDDL2.2 tasks to
multi-valued planning tasks, which we formally introduce in Section
\ref{section:definition}. The knowledge compilation component is
discussed in Section \ref{section:knowledge-compilation}, the search
component in Section \ref{section:search}. We conclude with a
presentation of experimental results in
Section~\ref{section:experiments} and some discussion in
Section~\ref{section:discussion}.

\section{Related Work}
\label{section:related-work}

As a planning system based on heuristic forward search, Fast Downward
is clearly related to other heuristic planners such as HSP \cite{hsp}
or FF \cite{ff} on the architectural level. However, in this section
we focus on work that is related on the \emph{conceptual} level, \ie,
work that uses similar forms of hierarchical decomposition of causal
graphs and work that uses similar forms of search in domain transition
graphs.

\subsection{Causal Graphs and Abstraction}
\label{section:related-work-abstractions}

The term \emph{causal graph} first appears in the literature in the
work by Williams and Nayak \citeyear{williams-nayak}, but the general
idea is considerably older. The approach of hierarchically
decomposing planning tasks is arguably as old as the field of AI
Planning itself, having first surfaced in Newell and Simon's
\citeyear{newell-simon:1963} work on the General Problem Solver.

Still, it took a long time for these notions to evolve to their
modern form. Sacerdoti's \citeyear{sacerdoti:aij74} ABSTRIPS algorithm
introduced the concept of \emph{abstraction spaces} for STRIPS-like
planning tasks. An abstraction space of a STRIPS task is the state
space of an \emph{abstracted task}, which is obtained by removing all
preconditions from the operators of the original task that belong to a
given set of propositions (which are \emph{abstracted
away}).\footnote{In later work by other authors, propositions which
are abstracted away are also removed from the operator effects. This
only makes a difference in subtle cases that require the presence of
axioms; we do not distinguish between these two kinds of abstraction
here.} To solve a planning task, ABSTRIPS first generates a plan for
an abstracted task, then refines this plan by inserting concrete plans
between the abstract plan steps that ``bridge the gap'' between
abstract states by satisfying the operator preconditions which were
ignored at the abstract level. The idea is easily generalized to
several levels of abstraction forming an \emph{abstraction hierarchy},
with a very abstract level at the top where almost all preconditions
are ignored, successively introducing more preconditions at every
layer until the final layer of the hierarchy equals the original
planning task.

One problem with this approach to planning is that in general there
is no guarantee that the abstract plans bear any resemblance to
reasonable concrete plans. For example, if abstraction spaces are
chosen badly, it is quite possible that finding a concrete plan that
satisfies the precondition of the first operator in the abstract plan
is more difficult than solving the original goal at the concrete
level. Such shortcomings spawned a large amount of research on the
properties of abstraction hierarchies and how they can be generated
automatically.

Tenenberg \citeyear{tenenberg:1991} gives one of the
first formal accounts of the properties of different kinds of
abstraction. Among other contributions, he defines the so-called
\emph{upward solution property}, which can be informally stated as:
``If there exists a concrete solution, then there also exists an
abstract solution''. Rather surprisingly, not all abstractions
considered at the time satisfied this very basic property, without
which one would be loathe to call a given state space an
``abstraction'' of another state space.

A limitation of the upward solution property is that it states no
relationship between the concrete and abstract plan at all. For
ABSTRIPS-style hierarchical planning to be successful, the abstract
plan must bear some resemblance to a concrete one; otherwise there is
little point in trying to refine it. Indeed, Tenenberg introduces
stronger versions of the upward solution property, but more relevant
to Fast Downward is Knoblock's \citeyear{knoblock-abstractions} work
on the \emph{ordered monotonicity property}. An abstraction space
satisfies the ordered monotonicity property if, roughly speaking, any
concrete solution can be derived from some abstract solution while
leaving the actions in the abstract plan intact and relevant to the
concrete plan. Clearly, this is a very important property for
ABSTRIPS-like hierarchical planning.

It is in Knoblock's article that causal graphs first surface (although
he does not introduce a name for them). Translated to our terminology,
Knoblock proves the following relationship between useful abstractions
and causal graphs: \emph{If the causal graph contains no path from a
variable that is not abstracted away to a variable that is abstracted
away, then the abstraction has the ordered monotonicity property.} In
particular, this means that for acyclic causal graphs, it is possible
to devise an abstraction hierarchy where only one new variable is
introduced at each level.

Besides these theoretical contributions, Knoblock presents a
planning system called ALPINE which computes an abstraction hierarchy
for a planning task from its causal graph and exploits this within a
hierarchical refinement planner. Although the planning method is very
different, the derivation of the abstraction hierarchy is very similar
to Fast Downward's method for generating hierarchical decompositions
of planning tasks (Section \ref{section:causal-graphs}).

By itself, the ordered monotonicity property is not sufficient to
guarantee good performance of a hierarchical planning approach. It
guarantees that every concrete solution can be obtained in a natural
way from an abstract solution, but it does not guarantee that all
abstract solutions can be refined to concrete ones. Such a guarantee
is provided by the \emph{downward refinement property}, introduced by
Bacchus and Yang \citeyear{bacchus-downward-refinement}.

The downward refinement property can rarely be guaranteed in actual
planning domains, so Bacchus and Yang develop an analytical model for
the performance of hierarchical planning in situations where a given
abstract plan can only be refined with a certain probability $p < 1$.
Based on this analysis, they present an extension to ALPINE called
HIGHPOINT, which selects an abstraction hierarchy with high refinement
probability among those that satisfy the ordered monotonicity
property. In practice, it is not feasible to compute the refinement
probability, so HIGHPOINT approximates this value based on the notion
of \emph{$k$-ary necessary connectivity}.

\subsection{Causal Graphs and Unary STRIPS Operators}

Causal graphs are first given a name by Jonsson and B\"ackstr\"om
\citeyear{jonsson-backstrom:ewsp95,jonsson-3s}, who call them
\emph{dependency graphs}. They study a fragment of propositional
STRIPS with negative conditions which has the interesting property
that plan existence can be decided in polynomial time, but minimal
solutions to a task can be exponentially long, so that no polynomial
planning algorithm exists. They present an \emph{incremental} planning
algorithm with polynomial delay, \ie, a planning algorithm that
decides within polynomial time whether or not a given task has a
solution, and, if so, generates such a solution step by step, requiring
only polynomial time between any two subsequent
steps.\footnote{However, there is no guarantee that the length of the
generated solution is polynomially related to the length of an
optimal solution; it might be exponentially longer. Therefore, the
algorithm might spend exponential time on tasks that can be solved
in polynomial time.}

The fragment of STRIPS covered by Jonsson and
B\"ackstr\"om's algorithm is called \emph{3S} and is defined by the
requirement that the causal graph of the task is acyclic and each
state variables is \emph{static}, \emph{symmetrically reversible}, or
\emph{splitting}. \emph{Static} variables are those for which it is
easy to guarantee that they never change their value in any solution
plan. These variables can be detected and compiled away easily.
\emph{Symmetrically reversible} variables are those where for each
operator which makes them true there is a corresponding operator with
identical preconditions which makes them false, and vice versa. In
other words, a variable is symmetrically reversible iff its domain
transition graph is undirected. Finally, a variable $v$ is
\emph{splitting} iff its removal from the causal graph weakly
disconnects its positive successors (those variables which appear in
effects of operators of which $v$ is a precondition) from its negative
successors (those variables which appear in effects of operators of
which $\neg v$ is a precondition).

Williams and Nayak \citeyear{williams-nayak} independently prove that
incremental (or, in their setting, \emph{reactive}) planning is a
polynomial problem in a STRIPS-like setting where causal graphs are
acyclic and all operators are reversible. If all operators are
reversible (according to the definition by Williams and Nayak), all
variables are symmetrically reversible (according to the definition by
Jonsson and B\"ackstr\"om), so this is actually a special case of the
previous result. However, Williams and Nayak's work applies to a more
general formalism than propositional STRIPS, so that the approaches are not
directly comparable.

More recently, Domshlak and Brafman provide a detailed account of the
complexity of finding plans in the propositional STRIPS (with
negation) formalism with unary operators and acyclic graphs
\cite{carmel-unary,brafman-domshlak:jair2003}.\footnote{According to
our formal definition of causal graphs in Section
\ref{section:causal-graphs}, operators with several effects always
induce cycles in the causal graph, so \emph{acyclic causal graph}
implies \emph{unary operators}. Some researchers define causal
graphs differently, so we name both properties explicitly here.}
Among other results, they prove that the restriction to unary
operators and acyclic graphs does not reduce the complexity of plan
existence: the problem is \PSPACE-complete, just like unrestricted
propositional STRIPS planning \cite{complexity-bylander}. They also
show that for singly connected causal graphs, shortest plans cannot be
exponentially long, but the problem is still \NP-complete. For an even
more restricted class of causal graphs, namely polytrees of bounded
indegree, they present a polynomial planning algorithm. More
generally, their analysis relates the complexity of STRIPS planning in
unary domains to the \emph{number of paths} in their causal graph.

\subsection{Multi-Valued Planning Tasks}

With the exception of Williams and Nayak's paper, all the work
discussed so far exclusively deals with \emph{propositional} planning
problems, where all state variables assume values from a binary
domain. As we observed in the introduction, the question of
propositional vs.~multi-valued encodings usually has a strong impact
on the connectivity of the causal graph of a task. In fact, apart
from the trivial {\movie} domain, none of the common planning
benchmarks exhibits an acyclic causal graph when considering its
propositional representation. By contrast, the multi-valued encoding
of our introductory example does have an acyclic causal graph.

Due to the dominance of the PDDL (and previously, STRIPS) formalism,
non-binary state variables are not studied very often in the classical
planning literature. One of the most important exceptions to this rule
is the work on the {\sasplus} planning formalism, of which the papers
by B\"ackstr\"om and Nebel \citeyear{complexity-sas} and Jonsson and
B\"ackstr\"om \citeyear{jonsson-sas} are most relevant to Fast Downward.
The {\sasplus} planning formalism is basically equivalent to the
\emph{multi-valued planning tasks} we introduce in Section
\ref{section:definition} apart from the fact that it does not include
derived variables (axioms) or conditional effects. B\"ackstr\"om and
Nebel analyse the complexity of various subclasses of the
{\sasplus} formalism and discover three properties (\emph{unariness},
\emph{post-uniqueness} and \emph{single-valuedness}) that together
allow optimal planning in polynomial time. One of these three
properties (unariness) is related to acyclicity of causal graphs, and
one (post-uniqueness) implies a particularly simple shape of domain
transition graphs (namely, in post-unique tasks, all domain transition
graphs must be simple cycles or trees).

B\"ackstr\"om and Nebel do not analyse domain transition graphs
formally. Indeed, the term is only introduced in the later article by
Jonsson and B\"ackstr\"om \citeyear{jonsson-sas}, which refines the
earlier results by introducing five additional restrictions for
{\sasplus} tasks, all of which are related to properties of
domain transition graphs.

Neither of these two articles discusses the notion of causal graphs.
Indeed, the only earlier work we are aware of which includes
\emph{both} causal graphs and domain transition graphs as central
concepts is the article by Domshlak and Dinitz \citeyear{carmel-ecp99}
on the \emph{state-transition support} (STS) problem, which is
essentially equivalent to {\sasplus} planning with unary operators. In
the context of STS, domain transition graphs are called \emph{strategy
  graphs} and causal graphs are called \emph{dependence graphs}, but
apart from minor details, the semantics of the two formalisms are
identical. Domshlak and Dinitz provide a map of the complexity of the
STS problem in terms of the shape of its causal graph, showing that
the problem is \NP-complete or worse for almost all non-trivial cases.
One interesting result is that if the causal graph is a simple chain
of $n$ nodes and all variables are three-valued, the length of minimal
plans can already grow as $\Omega(2^n)$. By contrast,
\emph{propositional} tasks with the same causal graph shape admit
polynomial planning algorithms according to the result by Brafman and
Domshlak \citeyear{brafman-domshlak:jair2003}, because such causal
graphs are polytrees with a constant indegree bound (namely, a bound
of $1$).

To summarize and conclude our discussion of related work, we observe
that the central concepts of Fast Downward and the causal graph
heuristic, such as causal graphs and domain transition graphs, are
firmly rooted in previous work. However, Fast Downward is the first
attempt to marry hierarchical problem decomposition to the use of
multi-valued state variables within a general planning framework. It
is also the first attempt to apply techniques similar to those of
Knoblock \citeyear{knoblock-abstractions} and Bacchus and Yang
\citeyear{bacchus-downward-refinement} within a heuristic search
planner.

The significance of this latter point should not be
underestimated: For classical approaches to hierarchical problem
decomposition, it is imperative that an abstraction satisfies the
ordered monotonicity property, and it is important that the
probability of being able to refine an abstract plan to a concrete
plan is high, as the analysis by Bacchus and Yang shows.
Unfortunately, non-trivial abstraction hierarchies are rarely
ordered monotonic, and even more rarely guarantee high
refinement probabilities. Within a heuristic approach, these
``must-haves'' turn into ``nice-to-haves'': If an abstraction
hierarchy is not ordered monotonic or if an abstract plan considered by
the heuristic evaluator is not refinable, this merely reduces the
quality of the heuristic estimate, rather than causing the search to
fail (in the worst case) or spend a long time trying to salvage
non-refinable abstract plans (in the not much better case).

\section{Fast Downward}
\label{section:architecture}

We will now describe the overall architecture of the planner. Fast
Downward is a classical planning system based on the ideas of
heuristic forward search and hierarchical problem decomposition. It
can deal with the full range of propositional PDDL2.2
\cite{pddl2,pddl2.2}, {\ie}, in addition to STRIPS planning, it
supports arbitrary formulae in operator preconditions and goal
conditions, and it can deal with conditional and universally
quantified effects and derived predicates (axioms).

The name of the planner derives from two sources: Of course, one of
these sources is Hoffmann's very successful FF (``Fast Forward'')
planner \cite{ff}. Like FF, Fast Downward is a heuristic progression
planner, {\ie}, it computes plans by heuristic search in the space of
world states reachable from the initial situation. However, compared
to FF, Fast Downward uses a very different heuristic evaluation
function called the \emph{causal graph heuristic}. The heuristic
evaluator proceeds ``downward'' in so far as it tries to solve
planning tasks in the hierarchical fashion outlined in the introduction.
Starting from top-level goals, the algorithm recurses further and
further down the causal graph until all remaining subproblems are
basic graph search tasks.

Similar to FF, the planner has shown excellent performance: The original
implementation of the causal graph heuristic, plugged into a standard
best-first search algorithm, outperformed the previous champions in that
area, FF and LPG \cite{lpg}, on the set of STRIPS benchmarks from the
first three international planning competitions \cite{malte-icaps-cg}.
Fast Downward itself followed in the footsteps of FF and LPG by winning
the propositional, non-optimizing track of the 4th International
Planning Competition at ICAPS 2004 (referred to as IPC4 from now on).

As mentioned in the introduction, Fast Downward solves a planning task
in three phases (Fig.~\ref{figure:modules}):

\includepsfigure{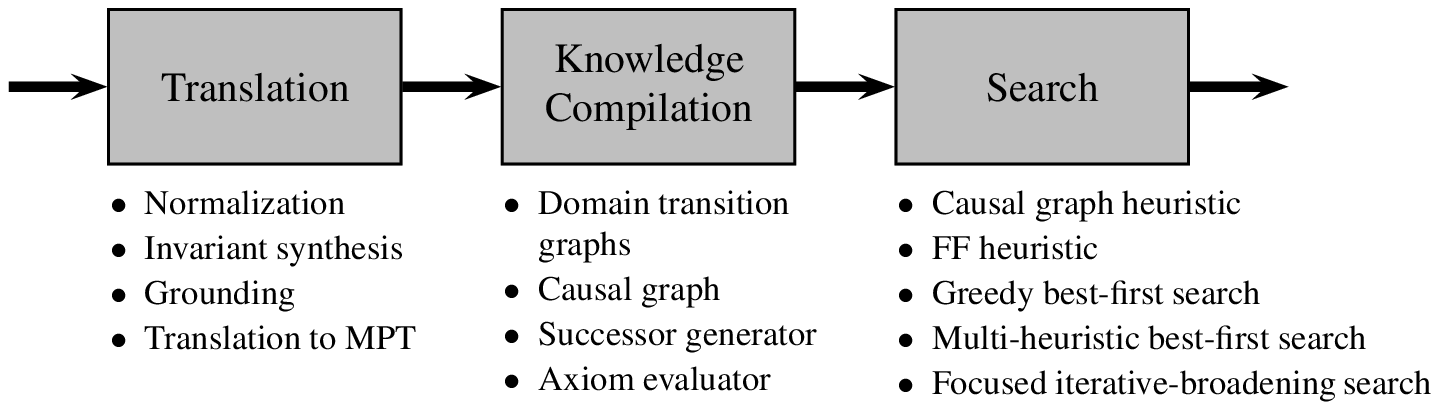}{The three phases of Fast Downward's
  execution.}

\begin{itemize}
\item The \emph{translation} component is responsible for transforming
  the PDDL2.2 input into a non-binary form which is more amenable to
  hierarchical planning approaches. It applies a number of
  normalizations to compile away syntactic constructs like
  disjunctions which are not directly supported by the causal graph
  heuristic and performs grounding of axioms and operators. Most
  importantly, it uses invariant synthesis methods to find groups of
  related propositions which can be encoded as a single multi-valued
  variable. The output of the translation component is a
  \emph{multi-valued planning task}, defined in the following section.
\item The \emph{knowledge compilation} component generates four kinds of
  data structures that play a central role during search: \emph{Domain
  transition graphs} encode how, and under what conditions, state
  variables can change their values. The \emph{causal graph} represents
  the hierarchical dependencies between the different state variables.
  The \emph{successor generator} is an efficient data structure for
  determining the set of applicable operators in a given state.
  Finally, the \emph{axiom evaluator} is an efficient data structure for
  computing the values of derived variables. The knowledge
  compilation component is described in Section
  \ref{section:knowledge-compilation}.
\item The \emph{search} component implements three different search
  algorithms to do the actual planning. Two of these algorithms make
  use of heuristic evaluation functions: One is the well-known greedy
  best-first search algorithm, using the causal graph heuristic. The
  other is called \emph{multi-heuristic best-first search}, a variant
  of greedy best-first search that tries to combine several heuristic
  evaluators in an orthogonal way; in the case of Fast Downward, it
  uses the causal graph and FF heuristics. The third search algorithm
  is called \emph{focused iterative-broadening search}; it is closely
  related to Ginsberg and Harvey's \citeyear{ginsberg-harvey:aij92}
  iterative broadening. It is not a heuristic search algorithm in the
  sense that it does not use an explicit heuristic evaluation
  function. Instead, it uses the information encoded in the causal
  graph to estimate the ``usefulness'' of operators towards satisfying
  the goals of the task. The search component is described in Section
  \ref{section:search}.
\end{itemize}

\section{Multi-Valued Planning Tasks}
\label{section:definition}

Let us now formally introduce the problem of planning with
multi-valued state variables. Our formalism is based on the
{\sasplus} planning model \cite{complexity-sas,jonsson-sas},
but extends it with axioms and conditional effects.

\begin{definition}{mpt}{Multi-valued planning tasks (MPTs)}
A \definedTerm{multi-valued planning task (MPT)} is given by a
5-tuple $\Pi = \langle \mathcal V, s_0, s_\star, 
\mathcal A, \mathcal O \rangle$ with the following components:
\begin{itemize}
\item
  $\mathcal V$ is a finite set of \definedTerm{state variables}, each
  with an associated finite domain $\mathcal D_v$.
  State variables are partitioned into \definedTerm{fluents} (affected by
  operators) and \definedTerm{derived variables} (computed by evaluating
  axioms). The domains of derived variables must contain the
  \definedTerm{undefined value} $\bot$.

  A \definedTerm{partial variable assignment} or \definedTerm{partial
  state} over $\mathcal V$ is a function $s$ on some subset of
  $\mathcal V$ such that $s(v) \in \mathcal D_v$ wherever $s(v)$ is
  defined. A partial state is called an \definedTerm{extended state}
  if it is defined for all variables in $\mathcal V$ and a
  \definedTerm{reduced state} or \definedTerm{state} if it is defined
  for all fluents in $\mathcal V$. In the context of partial variable
  assignments, we write $v=d$ for the variable-value pairing $(v, d)$
  or $v \mapsto d$.
\item
  $s_0$ is a state over $\mathcal V$ called the \definedTerm{initial state}.
\item
  $s_\star$ is a partial variable assignment over $\mathcal V$ called
  the \definedTerm{goal}.
\item
  $\mathcal A$ is a finite set of (MPT) \definedTerm{axioms} over
  $\mathcal V$. Axioms are triples of the form $\langle \cond, v, d
  \rangle$, where $\cond$ is a partial variable assignment called the
  \definedTerm{condition} or \definedTerm{body} of the axiom,
  $v$ is a derived variable called the \definedTerm{affected variable},
  and $d \in \mathcal D_v$ is called the \definedTerm{derived value}
  for $v$. The pair $(v, d)$ is called the
  \definedTerm{head} of the axiom and can be written as $v:=d$.

  The axiom set $\mathcal A$ is partitioned into a totally ordered set
  of \definedTerm{axiom layers} $\mathcal A_1 \prec \dots \prec \mathcal A_k$
  such that within the same layer, each affected variable may only be
  associated with a single value in axiom heads and bodies. In other
  words, within the same layer, axioms with the same affected variable
  but different derived values are forbidden, and if a variable
  appears in an axiom head, then it may not appear with a different
  value in a body. This is called the \definedTerm{layering property}.
\item
  $\mathcal O$ is a finite set of (MPT) \definedTerm{operators} over $\mathcal
  V$. An operator $\langle \pre, \eff \rangle$ consists of a partial
  variable assignment $\pre$ over $\mathcal V$ called its
  \definedTerm{precondition}, and a finite set of \definedTerm{effects} $\eff$.
  Effects are triples $\langle \cond, v, d \rangle$, where $\cond$ is a (possibly
  empty) partial variable assignment called the \definedTerm{effect condition},
  $v$ is a fluent called the \definedTerm{affected variable}, and $d \in
  \mathcal D_v$ is called the \definedTerm{new value} for $v$.
\end{itemize}

For axioms and effects, we also use the notation $\cond
\rightarrow v:=d$ in place of $\langle \cond, v, d \rangle$.
\end{definition}

To provide a formal semantics for MPT planning, we first need to
formalize axioms:

\begin{definition}{extended-state}{Extended states defined by a state}
Let $s$ be a state of an MPT $\Pi$ with axioms $\mathcal A$, layered
as $\mathcal A_1 \prec \dots \prec \mathcal A_k$. The \definedTerm{extended
state defined by $s$}, written as $\mathcal A(s)$, is the result $s'$
of the following algorithm:

\smallskip

\begin{algorithm}
  \hspace*{0cm}\textbf{algorithm} evaluate-axioms($\mathcal A_1$, $\dots$,
    $\mathcal A_k$, $s$):\\
  \hspace*{1cm}\textbf{for each} variable $v$: \\
  \hspace*{2cm}$s'(v) := \begin{cases}
    s(v) & \text{if $v$ is a fluent variable} \\
    \bot & \text{if $v$ is a derived variable}\end{cases}$ \\
  \hspace*{1cm}\textbf{for} $i \in \{1, \dots, k\}$: \\
  \hspace*{2cm}\textbf{while} there exists an axiom
    $(\cond \rightarrow v := d) \in \mathcal A_i$ with
    $\cond \subseteq s'$ \textbf{and} $s'(v) \neq d$: \\
  \hspace*{3cm}Choose such an axiom $\cond \rightarrow v := d$. \\
  \hspace*{3cm}$s'(v) := d$
\end{algorithm}
\end{definition}

In other words, axioms are evaluated in a layer-by-layer fashion using
fixed point computations, which is very similar to the semantics of
stratified logic programs. It is easy to see that the layering
property from Definition \ref{def:mpt} guarantees that the algorithm
terminates and produces a deterministic result. Having defined the
semantics of axioms, we can now define the state space of an MPT:

\begin{definition}{mpt-state-space}{MPT state spaces}
The \definedTerm{state space} of an MPT
$\Pi = \langle \mathcal V, s_0, s_\star, \mathcal A, \mathcal O \rangle$,
denoted as $\mathcal S(\Pi)$, is a directed graph. Its vertex set is
the set of states of $\mathcal V$, and it contains an arc $(s,s')$ iff
there exists some operator $\langle \pre, \eff \rangle \in \mathcal O$
such that:
\begin{itemize}
\item $\pre \subseteq \mathcal A(s)$,
\item $s'(v) = d$ for all effects $\cond \rightarrow v := d \in
  \eff$ such that $\cond \subseteq \mathcal A(s)$, and
\item $s'(v) = s(v)$ for all other fluents.
\end{itemize}
\end{definition}

Finally, we can define the MPT planning problem:
\begin{definition}{mpt-planning-problem}{MPT planning}
{\mptplanex} is the following decision problem: Given an MPT $\Pi$
with initial state $s_0$ and goal $s_\star$, does
$\mathcal S(\Pi)$ contain a path from $s_0$ to some state $s'$ with
$s_\star \subseteq \mathcal A(s')$?

\smallskip

\noindent
{\mptplan} is the following search problem: Given an MPT $\Pi$
with initial state $s_0$ and goal $s_\star$, compute a path in
$\mathcal S(\Pi)$ from $s_0$ to some state $s'$ with $s_\star
\subseteq \mathcal A(s')$, or prove that none exists.
\end{definition}

The {\mptplanex} problem is easily shown to be \PSPACE-hard because it
generalizes the plan existence problem for propositional STRIPS, which
is known to be \PSPACE-complete \cite{complexity-bylander}. It is also
easy to see that the addition of multi-valued domains, axioms and
conditional effects does not increase the theoretical complexity of MPT
planning beyond propositional STRIPS. Thus, we conclude our formal
introduction of MPT planning by stating that {\mptplanex} is
\PSPACE-complete, and turn to the practical side of things in the
following section.

\section{Knowledge Compilation}
\label{section:knowledge-compilation}

The purpose of the knowledge compilation component is to set the stage
for the search algorithms by compiling the critical information about
the planning task into a number of data structures for efficient
access. In other contexts, computations of this kind are often called
\emph{preprocessing}. However, ``preprocessing'' is such a nondescript
word that it can mean basically anything. For this reason, we prefer a term that
puts a stronger emphasis on the role of this module: To rephrase the
critical information about the planning task in such a way that it is
directly useful to the search algorithms. Of the three building blocks
of Fast Downward (translation, knowledge compilation, search), it is
the least time-critical part, always requiring less time than
translation and being dominated by search for all but the most trivial
tasks.

Knowledge compilation comprises three items. First and foremost, we
compute the \emph{domain transition graph} of each state variable. The
domain transition graph for a state variable encodes under what
circumstances that variable can change its value, {\ie}, from which
values in the domain there are transitions to which other values,
which operators or axioms are responsible for the transition, and
which conditions on other state variables are associated with the
transition. Domain transition graphs are described in Section
\ref{section:domain-transition-graphs}. They are a central concept for
the computation of the causal graph heuristic, described in Section
\ref{section:cg-heuristic}.

Second, we compute the \emph{causal graph} of the planning task. Where
domain transition graphs encode dependencies between values for a
given state variable, the causal graph encodes dependencies between
different state variables. For example, if a given location in a
planning task can be unlocked by means of a key that can be carried by
the agent, then the variable representing the lock state of the
location is dependent on the variable that represents whether or not
the key is being carried. This dependency is encoded as an arc in the
causal graph. Like domain transition graphs, causal graphs are a
central concept for the computation of the causal graph heuristic, giving it its
name. The causal graph heuristic requires causal graphs to be acyclic. For this
reason, the knowledge compilation component also generates an acyclic
subgraph of the real causal graph when cycles occur. This amounts to a
relaxation of the planning task where some operator preconditions are
ignored. In addition to their usefulness for the causal graph heuristic, causal
graphs are also a key concept of the \emph{focused iterative-broadening
search} algorithm introduced in Section
\ref{section:focused-iterative-broadening-search}. We discuss causal
graphs in Section \ref{section:causal-graphs}.

Third, we compute two data structures that are useful for any
forward-searching algorithm for MPTs, called \emph{successor
generators} and \emph{axiom evaluators}. Successor generators
compute the set of applicable operators in a given world state, and
axiom evaluators compute the values of derived variables for a given
reduced state. Both are designed to do their job as
quickly as possible, which is especially important for the focused
iterative-broadening search algorithm, which does not compute
heuristic estimates and thus requires the basic operations for
expanding a search node to be implemented efficiently. These data
structures are discussed in Section
\ref{section:successor-generators}.

\subsection{Domain Transition Graphs}
\label{section:domain-transition-graphs}

The domain transition graph of a state variable is a representation of
the ways in which the variable can change its value, and of the
conditions that must be satisfied for such value changes to be
allowed. Domain transition graphs were introduced by Jonsson and
B\"ackstr\"om \citeyear{jonsson-sas} in the context of {\sasplus}
planning. Our formalization of domain transition graphs
generalizes the original definition to planning tasks involving axioms
and conditional effects.

\begin{definition}{domain-transition-graph}{Domain transition graphs}
Let $\Pi = \langle \mathcal V, s_0, s_\star, \mathcal A, \mathcal O
\rangle$ be a multi-valued planning task, and let $v \in \mathcal V$
be a state variable of $\Pi$.

The \definedTerm{domain transition graph} of $v$, in symbols
$\textit{DTG}(v)$, is a labelled directed graph with vertex set
$\mathcal D_v$.
If $v$ is a fluent, $\textit{DTG}(v)$ contains the following arcs:
\begin{itemize}
\item For each effect $\cond \rightarrow v := d'$ of an
  operator $o$ with precondition $\pre$ such that $\pre \cup \cond$
  contains some condition $v = d$, an arc from $d$ to $d'$ labelled
  with $\pre \cup \cond \setminus \{v=d\}$.
\item For each effect $\cond \rightarrow v := d'$ of an
  operator $o$ with precondition $\pre$ such that $\pre \cup \cond$
  does not contain the condition $v = d$ for any $d \in \mathcal D_v$,
  an arc from each $d \in \mathcal D_v \setminus \{d'\}$ to $d'$
  labelled with $\pre \cup \cond$.
\end{itemize}

If $v$ is a derived variable, $\textit{DTG}(v)$ contains the following
arcs:
\begin{itemize}
\item For each axiom $\cond \rightarrow v := d' \in \mathcal A$ such
  that $\cond$ contains some condition $v = d$, an arc from $d$ to
  $d'$ labelled with $\cond \setminus \{v=d\}$.
\item For each axiom $\cond \rightarrow v := d' \in \mathcal A$ such
  that $\cond$ does not contain the condition $v = d$ for any $d \in
  \mathcal D_v$, an arc from each $d \in \mathcal D_v \setminus
  \{d'\}$ to $d'$ labelled with $\cond$.
\end{itemize}

Arcs of domain transition graphs are called \definedTerm{transitions}.
Their labels are referred to as the \definedTerm{conditions} of the
transition.

Domain transition graphs can be weighted, in which case each
transition has an associated non-negative integer weight. Unless
stated otherwise, we assume that all transitions derived from
operators have weight 1 and all transitions derived from axioms have
weight 0.
\end{definition}

The definition is somewhat lengthy, but its informal content is easy
to grasp: The domain transition graph for $v$ contains a transition
from $d$ to $d'$ if there exists some operator or axiom that can
change the value of $v$ from $d$ to $d'$. Such a transition is
labelled with the conditions on \emph{other} state variables that must
be true if the transition shall be applied. Multiple transitions
between the same values using different conditions are allowed and
occur frequently.

\includepsfigure{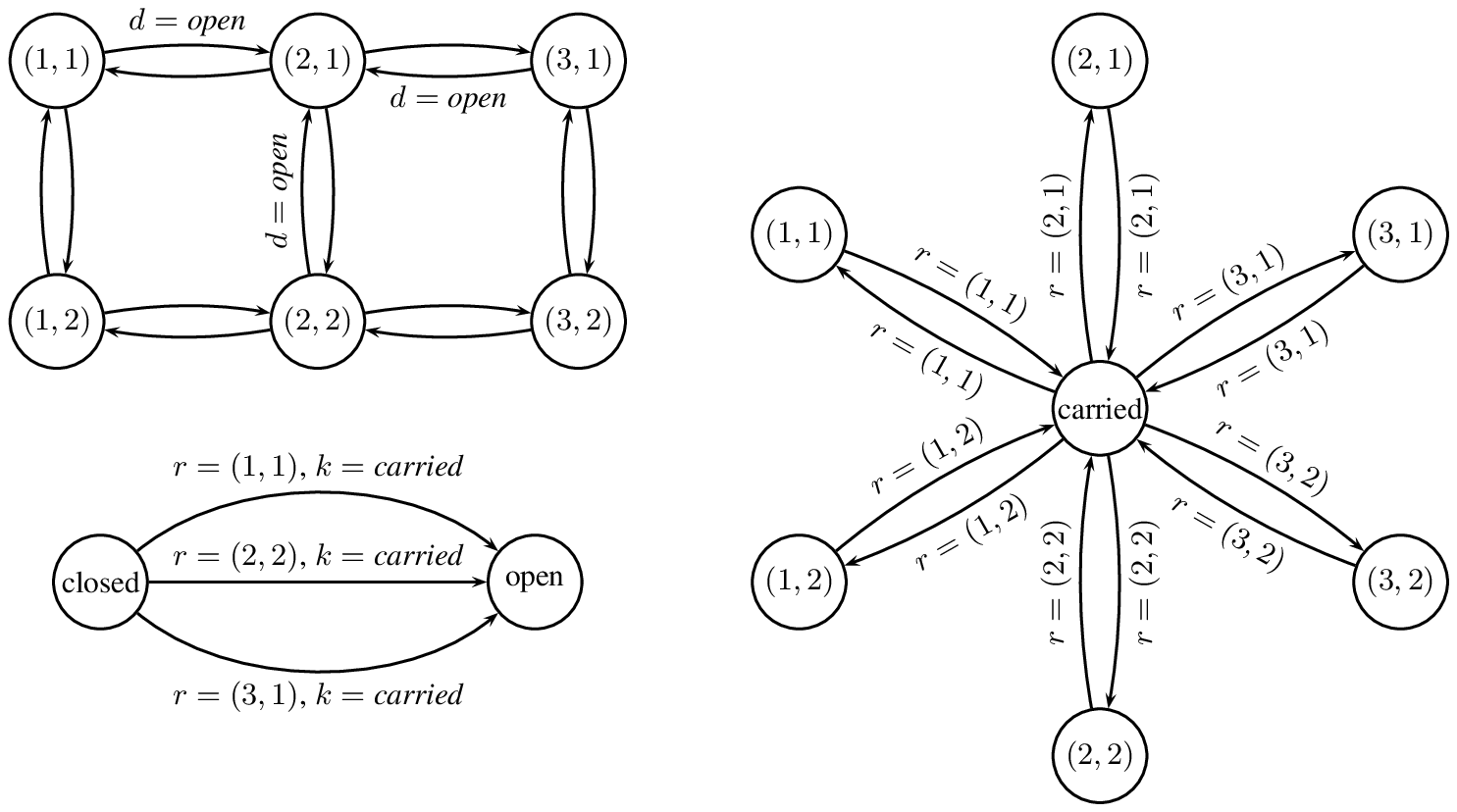}{Domain transition
  graphs of a {\grid} task.
  Top left: $\textit{DTG}(r)$
  (robot); right: $\textit{DTG}(k)$ (key); bottom left:
  $\textit{DTG}(d)$ (door).}

We have already seen domain transition graphs in the introductory
section (Figs.~\ref{figure:initial-example-parcel-dtg} and
\ref{figure:initial-example-other-dtgs}), although they were only
introduced informally and did not show the arc labels usually
associated with transitions.
Fig.~\ref{figure:domain-transition-graphs-grid} shows some examples
from a simple task in the {\grid} domain, featuring a $3\times2$ grid
with a single initially locked location in the centre of the upper
row, unlockable by a single key. In the MPT encoding of the task,
there are three state variables: variable $r$ with $\mathcal D_r =
\{~(x,y)~|~x \in \{1, 2, 3\}, \ y \in \{1, 2\}~\}$ encodes the
location of the robot, variable $k$ with $\mathcal D_k = \mathcal D_r
\cup \{\textit{carried}\}$ encodes the state of the key, and
variable $d$ with $\mathcal D_d = \{\textit{closed},
\textit{open}\}$ encodes the state of the initially locked grid
location.

If all operators of an MPT are unary ({\ie}, only have a single
effect) and we leave aside axioms for a moment, then there is a strong
correspondence between the state space of an MPT and its domain
transition graphs. Since vertices in domain transition graphs
correspond to values of state variables, a given state is represented
by selecting one vertex in each domain transition graph, called the
\emph{active vertex} of this state variable. Applying an operator
means changing the active vertex of some state variable by performing
a transition in the corresponding domain transition graph. Whether or
not such a transition is allowed depends on its condition, which is
checked against the active vertices of the other domain transition
graphs.

Let us use the {\grid} example to illustrate this correspondence.
Consider an initial state where the robot is at location $(1,1)$, the
key is at location $(3,2)$, and the door is locked. We represent this
by placing pebbles on the appropriate vertices of the three domain
transition graphs. We want to move the pebble in the domain transition
graph of the key to location $(2,1)$. This can be done by moving the
robot pebble to vertex $(1,2)$, then $(2,2)$, then $(3,2)$, moving the
key pebble to the vertex \emph{carried}, moving the robot pebble back
to vertex $(2,2)$, moving the door pebble to \emph{open}, moving the
robot pebble to vertex $(2,1)$ and finally moving the key pebble to
vertex $(2,1)$.

The example shows how plan execution can be viewed as simultaneous
traversal of domain transition graphs \cite<cf.>{carmel-ecp99}. This
is an important notion for Fast Downward because the causal graph heuristic
computes its heuristic estimates by solving subproblems of the
planning task by looking for paths in domain transition graphs in
basically the way we have described.

As mentioned before, this view of MPT planning is only completely
accurate for unary tasks without axioms, for which the domain
transition graphs are indeed a complete representation of the state
space. For non-unary operators, we would need to ``link'' certain
transitions in different domain transition graphs which belong to the
same operator. These could then only be executed together. For axioms,
we would need to mark certain transitions as ``mandatory'', requiring
that they be taken whenever possible. (This is only intended as a
rough analogy and leaves out details like layered axioms.)

In our previous work \cite{malte-icaps-cg}, we have successfully
applied this view of planning to STRIPS tasks. Extending the notion to
plans with conditional effects provides no challenges because domain
transition graphs always consider planning operators one effect at a
time, in which case effect condition can simply be seen as part of the
operator precondition. However, axioms provide a challenge that is
easily overlooked. If we want to change the value of a fluent
from $d$ to $d'$, the domain transition graph contains
all the important information; just find a path from $d$ to $d'$ and
try to find out how the associated conditions can be achieved.
Consider the same problem for a derived state variable. Let us assume
that unlocking the location in the {\grid} example leads to a drought,
causing the robot to freeze if it enters a horizontally adjacent
location. We could encode this with a new derived variable $f$ (for
\emph{freezing}) with domain $\mathcal D_f = \{\top, \bot\}$, defined
by the axioms $d = \textit{open}, r = (1,1) \rightarrow f := \top$ and
$d = \textit{open}, r = (3,1) \rightarrow f := \top$. The domain
transition graph $\textit{DTG}(f)$ is depicted in
Fig.~\ref{figure:domain-transition-graph-grid-axiom} (left).

\includepsfigure{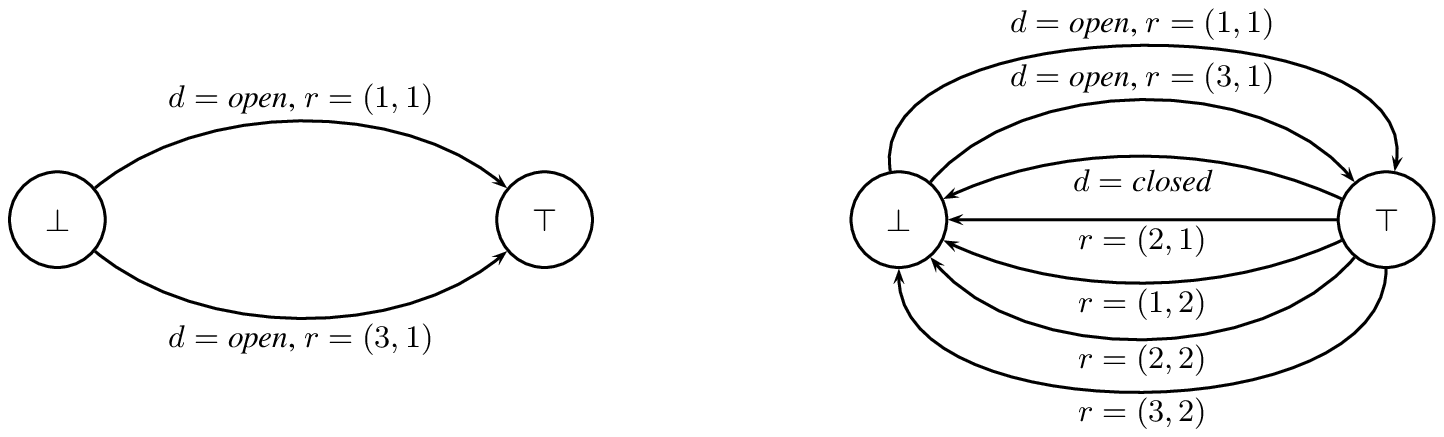}{Domain transition
  graphs for the \emph{freezing} variable in the {\grid} task, normal
  (left) and extended (right). Note that only the extended graph shows
  how to change state from ``freezing'' ($\top$) to ``not freezing''
  ($\bot$).}

The problem with that domain transition graph is that it does not tell
us how we can change the state of variable $f$ from $\top$ to $\bot$.
In general, in MPTs derived from STRIPS tasks where derived predicates
occur negatively in any condition, the domain transition graph does
not contain sufficient information for changing the value of a derived
variable from ``true'' to ``false''. Derived variables never assume
the value $\bot$ due to a \emph{derivation} of this value; because of
negation as failure semantics, they only assume the value \emph{by
default} if \emph{no other value} can be derived. If we want to
reason about ways of setting the value of a derived variable to
$\bot$, we will need to make this information explicit.

In logical notation, whether or not a derived variable assumes a given
value by triggering an axiom at a given layer is determined by a
formula in disjunctive normal form, with one disjunct for each axiom
setting the value. For example, our axioms  $d = \textit{open}, r =
(1,1) \rightarrow f := \top$ and $d = \textit{open}, r = (3,1)
\rightarrow f := \top$ correspond to the DNF formula
$(d = \textit{open} \land r = (1,1)) \lor (d = \textit{open} \land
r = (3,1))$. If we want to know when these rules do \emph{not}
trigger, we must negate this formula, leading to the CNF formula
$(d \neq \textit{open} \lor r \neq (1,1)) \land (d \neq \textit{open}
\lor r \neq (3,1))$. To be able to encode this information in the
domain transition graph, we need to replace the inequalities with
equalities and translate the formula back to DNF. Since such
transformations can increase the formula size dramatically, we apply
simplifications along the way, removing duplicated and dominated
disjuncts. The result in this case is the DNF formula
$d = \textit{closed} \lor r = (2, 1) \lor
r = (1, 2) \lor r = (2, 2) \lor r = (3, 2)$.

A domain transition graph for a derived variable which has been
enriched to contain the possible ways of causing the variable to
assume the value $\bot$ is called an \emph{extended domain transition
graph}, as shown for the {\grid} example in
Fig.~\ref{figure:domain-transition-graph-grid-axiom} (right). Since
computing the extended domain transition graph can be costly and is
not always necessary, the knowledge compilation component scans the
conditions of the planning task (axioms, operator preconditions and
effect conditions, goal) for occurrences of pairings of the type $v =
\bot$ for derived variables $v$. Extended domain transition graphs are
only computed for those derived variables for which they are required.

Note that negative occurrences of derived variables can cascade: If
$u$, $v$ and $w$ are derived variables with domain $\{\top, \bot\}$
and the condition $v = \bot$ is present in some operator precondition,
and moreover $v$ is defined by the axiom $u = \top, w = \top
\rightarrow v := \top$, then $v$ assumes the value $\bot$ whenever $u$
or $w$ do, so we would require extended domain transition graphs for
$u$ and $w$ as well.

On the other hand, multiple layers of negation as failure can cancel
each other out: If derived variable $v$ only occurs in conditions of
the form $v = \bot$ but never in positive form and is defined by the
axiom $u = \bot, w = \bot \rightarrow v := \top$, then we do not
necessarily require extended domain transition graphs for $u$ and $w$.

In general, whether or not we need extended domain transition graphs
for a derived variable is determined by the following rules:
\begin{itemize}
\item If $v$ is a derived variable for which the condition $v = d$
  for $d \neq \bot$ appears in an operator precondition, effect
  condition or in the goal, then $v$ \emph{is used positively}.
\item If $v$ is a derived variable for which the condition $v = \bot$
  appears in an operator precondition, effect condition or in the
  goal, then $v$ \emph{is used negatively}.
\item If $v$ is a derived variable for which the condition $v = d$
  for $d \neq \bot$ appears in the body of an axiom whose head is used
  positively (negatively), then $v$ is used positively (negatively).
\item If $v$ is a derived variable for which the condition $v = \bot$
  appears in the body of an axiom whose head is used
  positively (negatively), then $v$ is used negatively (positively).
\end{itemize}

The knowledge compilation component computes extended domain
transition graphs for all derived variables which are used negatively
and (standard) domain transition graphs for all other state variables.
Normal domain transition graphs are computed by going through the set
of axioms and the set of operator effects following Definition
\ref{def:domain-transition-graph}, which is reasonably
straight-forward; the computation of extended domain transition
graphs has been outlined above. Therefore, the algorithmic aspects of
this topic should not require further discussion.

\subsection{Causal Graphs}
\label{section:causal-graphs}

Causal graphs have been introduced informally in the introduction.
Here is a formal definition.

\begin{definition}{causal-graph}{Causal graphs}
Let $\Pi$ be a multi-valued planning task with variable set $\mathcal
V$. The \definedTerm{causal graph} of $\Pi$, in symbols
$\textit{CG}(\Pi)$, is the directed graph with vertex set $\mathcal V$
containing an arc $(v,v')$ iff $v \neq v'$ and one of the following
conditions is true:
\begin{itemize}
\item The domain transition graph of $v'$ has a transition with some
  condition on $v$.
\item The set of affected variables in the effect list of some
  operator includes both $v$ and $v'$.
\end{itemize}
In the first case, we say that an arc is induced by a \definedTerm{transition
condition}. In the second case we say that it is induced by
\definedTerm{co-occurring effects}.
\end{definition}

Of course, arcs induced by transition conditions and arcs induced by
co-occurring effects are not mutually exclusive. The same causal graph
arc can be generated for both reasons.

Informally, the causal graph contains an arc from a source variable to
a target variable if changes in the value of the target variable can
depend on the value of the source variable. Such arcs are included
also if this dependency is of the form of an \emph{effect} on the
source variable. This agrees with the definition of \emph{dependency
graphs} by Jonsson and B\"ackstr\"om \citeyear{jonsson-3s}, although
these authors distinguish between the two different ways in which an
arc in the graph can be introduced by using labelled arcs.

Whether or not co-occurring effects should induce arcs in the causal
graph depends on the intended semantics: If such arcs are not included, the
set of causal graph ancestors $\textit{anc}(v)$ of a variable $v$ are
precisely those variables which are relevant if our goal is to change
the value of $v$. Plans for this goal can be computed without
considering any variables outside $\textit{anc}(v)$, by eliminating
all variables outside $\textit{anc}(v)$ from the planning task and
simplifying axioms and operators accordingly. We call this the
\emph{achievability definition} of causal graphs, because causal
graphs encode what variables are important for achieving a given
assignment to a state variable.

However, with the achievability definition, a planner that only
considers $\textit{anc}(v)$ while generating an action sequence that
achieves a given valuation for $v$ may modify variables outside of
$\textit{anc}(v)$, {\ie}, the generated plans have side effects which
could destroy previously achieved goals or otherwise have a negative
impact on overall planning. Therefore, we prefer our definition, which
we call the \emph{separability definition} of causal graphs.

\subsubsection{Acyclic Causal Graphs}

Following the separability definition of causal graphs, solving a
subproblem over variables $\textit{anc}(v)$ is always possible without
changing any values outside of $\textit{anc}(v)$. This leads us to the
following observation.

\begin{observation}{easy-problems}{Acyclic causal graphs and
    strongly connected domain transition graphs}
Let $\Pi$ be an MPT such that $\textit{CG}(\Pi)$ is acyclic, all
domain transition graphs are strongly connected, there are no derived
variables, and no trivially false conditions occur in operators or
goals. Then $\Pi$ has a solution.
\end{observation}

By \emph{trivially false} conditions, we mean conditions of the kind
$\{v = d, v = d'\}$ for $d \neq d'$. Note the similarity of
Observation \ref{observation:easy-problems} to the results of
Williams and Nayak \citeyear{williams-nayak} on planning in domains
with unary operators, acyclic causal graphs and reversible
transitions. Under the separability definition of causal graphs,
acyclic causal graphs imply unariness of operators because operators
with several effects introduce causal cycles. Moreover, strong
connectedness of domain transition graphs is closely related to
Williams' and Nayak's reversibility property, although it is a weaker
requirement.

The truth of the observation can easily be seen inductively: If the
planning task has only one state variable and the domain transition
graph is strongly connected, then any state (of the one variable) can
be transformed into any other state by applying graph search
techniques. If the planning task has several state variables and the
causal graph is acyclic, we pick a sink of the causal graph, {\ie}, a
variable $v$ without outgoing arcs, and check if a goal is defined for
this variable. If not, we remove the variable from the task, thus
reducing the problem to one with fewer state variables, solved
recursively. If yes, we search for a path from $s_0(v)$ to
$s_\star(v)$ in the domain transition graph of $v$, which is
guaranteed to exist because the graph is strongly connected. This
yields a ``high-level plan'' for setting $v$ to $s_\star(v)$ which can
be fleshed out by recursively inserting the plans for setting the
variables of the predecessors of $v$ in the causal graph to the values
required for the transitions that form the high-level plan. Once the
desired value of $v$ has been set, $v$ can be eliminated from the
planning task and the remaining problem can be solved recursively.

  \begin{figure}[tbp]
    \begin{center}
      \input{figures/planning-algorithm-easy-task}
    \end{center}
    \caption{Planning algorithm for
  MPTs with acyclic causal graph and strongly connected domain
  transition graphs.}
    \label{figure:planning-algorithm-easy-task}
  \end{figure}

The algorithm is shown in
Fig.~\ref{figure:planning-algorithm-easy-task}. Although it is
backtrack-free, it can require exponential time to execute because the
generated plans can be exponentially long. This is unavoidable; even
for MPTs that satisfy the conditions of Observation
\ref{observation:easy-problems}, shortest plans can be exponentially
long. A family of planning tasks with this property is given in the proof of
Theorem 4.4 in the article by B\"ackstr\"om and Nebel
\citeyear{complexity-sas}.

This method for solving multi-valued planning tasks is
essentially \emph{planning by refinement}: We begin by constructing a
very abstract skeleton plan, which is merely a path in some domain
transition graph, then lower the level of abstraction by adding
operators to satisfy the preconditions required for the transitions
taken by the path. Strong connectedness of domain transition graphs
guarantees that every abstract plan can actually be refined to a
concrete plan. This is precisely Bacchus and Yang's
\citeyear{bacchus-downward-refinement} \emph{downward 
refinement property} (cf.~Section
\ref{section:related-work-abstractions}).

\subsubsection{Generating and Pruning Causal Graphs}

The usefulness of causal graphs for planning by refinement is not
limited to the acyclic case. Consider a subset $\mathcal V'$ of the
task variables which contains all its causal graph descendants. In
general, if we restrict the task to $\mathcal V'$ by removing all
occurrences of other variables from the initial state, goal, operators
and axioms, we obtain an abstraction of the original problem which
satisfies Knoblock's \citeyear{knoblock-abstractions} ordered
monotonicity property (Section
\ref{section:related-work-abstractions}).

Unfortunately, one major problem with this approach is that the
requirement to include all causal graph descendants is quite limiting.
It is not uncommon for the causal graph of a planning task to be
strongly connected, in which case this technique will not allow us to
abstract away any variables at all. However, in a heuristic approach,
we are free to simplify the planning task. In particular, by ignoring
some operator preconditions for the purposes of heuristic evaluation,
we can make an arbitrary causal graph acyclic. Clearly, the more
aspects of the real task we ignore, the worse we can expect our
heuristic to approximate the actual goal distance. Considering this, our aim is to
ignore as little information as possible. We will now explain how this
is done.

The knowledge compilation component begins its causal graph processing
by generating the ``full'' causal graph (Definition
\ref{def:causal-graph}). One consequence of the separability
definition of causal graphs is that all state variables which are not
ancestors of variables mentioned in the goal are completely
irrelevant. Therefore, having computed the graph, we then compute the
causal graph ancestors of all variables in the goal. Any state
variables which are not found to be goal ancestors are eliminated from
the planning task and causal graph, and associated operators and axioms
are removed.\footnote{This simplification is closely related to Knoblock's
  criterion for the \emph{problem-specific} ordered monotonicity
  property \cite{knoblock-abstractions}.}  Afterwards, we compute a
\emph{pruned causal graph}, an acyclic subgraph of the causal graph
with the same vertex set. We try do this in such a fashion that
``important'' causal dependencies are retained whenever possible. More
specifically, we apply the following algorithm.

First, we compute the strongly connected components of the causal
graph. Cycles only occur within strongly connected components, so each
component can be dealt with separately. Second, for each connected
component, we compute a total order $\prec$ on the vertices, retaining
only those arcs $(v,v')$ for which $v \prec v'$. If $v \prec v'$, we
say that $v'$ has a \emph{higher level} than $v$. The total order is
computed in the following way:

\begin{enumerate}
\item
  We assign a weight to each arc in the causal graph. The weight of an
  arc is $n$ if it is induced by $n$ axioms or operators. The
  lower the cumulated weight of the incoming arcs of a vertex, the
  fewer conditions are ignored by assigning a low level to this
  vertex.
\item 
  We then pick a vertex $v$ with minimal cumulated weight of
  incoming arcs and select it for the lowest level, {\ie}, we set
  $v \prec v'$ for all other vertices $v'$ in the strongly connected
  component.
\item Since $v$ has been dealt with, we remove the vertex and its
  incident arcs from consideration for the rest of the ordering
  algorithm.
\item The remaining problem is solved by iteratively applying the same
  technique to order the other vertices until only a single vertex
  remains.
\end{enumerate}

The reader will notice that the pruning choices within a strongly
connected component are performed by a greedy algorithm. We could also
try to find sets of arcs of minimal total weight such that
eliminating these arcs results in an acyclic graph. However, this is
an \NP-equivalent problem, even in the case of unweighted graphs
\cite[problem GT8]{garey-johnson}.

After generating the pruned causal graph, we also prune the domain
transition graphs by removing from the transition labels of
$\textit{DTG}(v)$ all conditions on variables $v'$ with $v \prec v'$.
These are the conditions that are ignored by the heuristic computation.
Finally, we simplify the domain transition graphs by removing
\emph{dominated transitions}: If $t$ and $t'$ are transitions between
the same two values of a variable, and the condition of $t$ is a
proper subset of the condition of $t'$, then transition $t$ is easier
to apply than $t'$, so that we remove $t'$.  Similarly, if there are
several transitions with identical conditions, we only keep one of
them.

\subsubsection{Causal Graph Examples}
\label{section:causal-graph-examples}

\includepsfigure{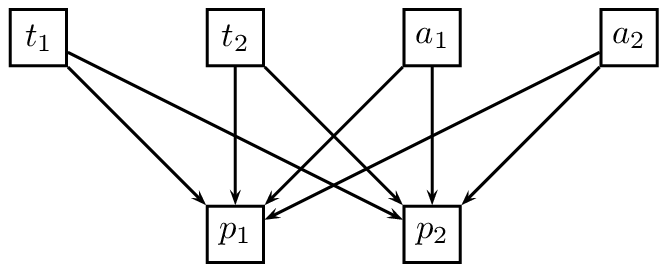}{Causal graph of a
  {\logistics} task. State variables $t_i$ and $a_i$ encode the
  locations of trucks and airplanes, state variables $p_i$ the
  locations of packages.}

To give some impression of the types of causal graphs typically found
in the standard benchmarks and the effects of pruning, we show some
examples of increasing graph complexity.

As our first and simplest example,
Fig.~\ref{figure:causal-graph-logistics} shows the causal graph of a
task from the {\logistics} domain, featuring two trucks, two airplanes
and two packages. As can be seen, the graph is acyclic, so it requires
no pruning for the causal graph heuristic. Since {\logistics} tasks
also feature strongly connected domain transition graphs, they can
even be solved by the polynomial \emph{solve-easy-MPT} algorithm.

\includepsfigure{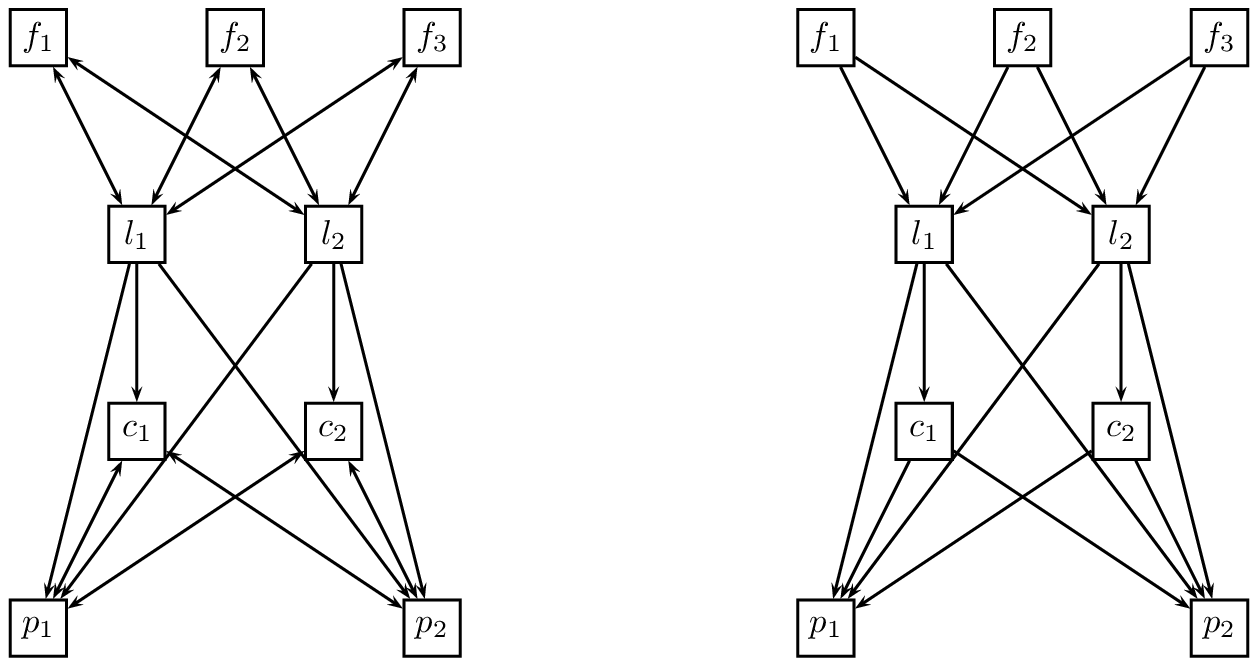}{Causal graph of a
  {\mystery} task (left) and of a relaxed version of the task (right).
  State variables $f_i$ encode the fuel at a location, state variables $l_i$
  and $c_i$ encode the locations and remaining capacities of trucks, and
  state variables $p_i$ encode the locations of packages.}

As a slightly more complicated example, the next figure,
Fig.~\ref{figure:causal-graph-mystery}, shows a task from the
{\mystery} domain with three locations, two trucks and two packages.
The causal graph contains a number of cycles, but these are mostly
local. By pruning arcs from vertices $l_i$ to $f_j$, we ignore the
fact that we must move trucks to certain locations if we want to use
up fuel at that location. As using up fuel is not a very useful thing
to do, this is not a big loss in information. By pruning arcs from
vertices $p_i$ to $c_j$, we ignore the fact that vehicles can only
increase or decrease their current capacity by unloading or loading
packages. Compared to heuristics based on ignoring delete effects,
this is not a great loss in information, since ignoring delete effects
in the {\mystery} domain almost amounts to ignoring capacity and fuel
constraints altogether. By pruning just these arcs, we can eliminate
all cycles in the causal graph, so the {\mystery} domain can be
considered fairly well-behaved.

\includepsfigure{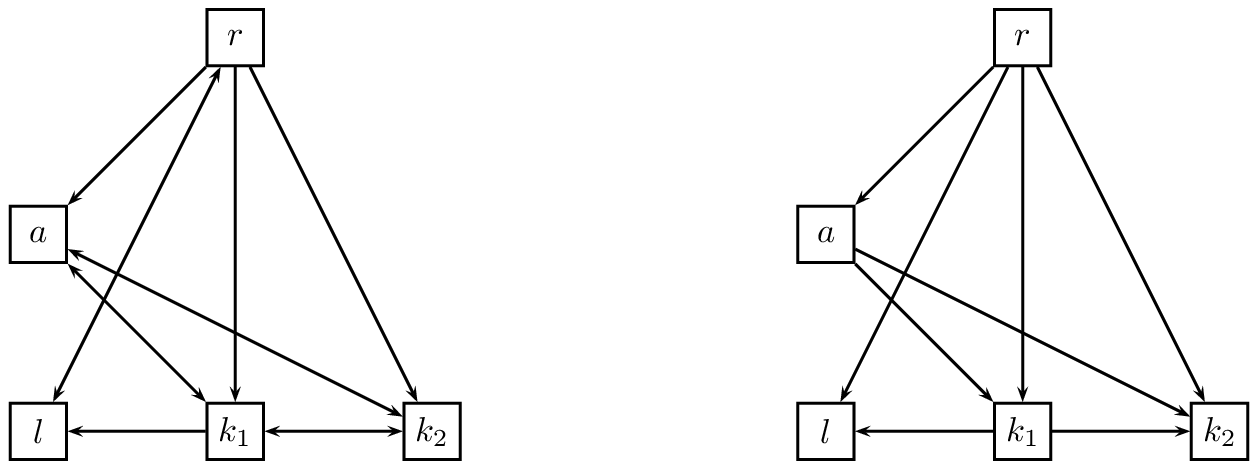}{Causal graph of a {\grid} task
  (left) and of a relaxed version of the task (right). State variable
  $r$ encodes the location of the robot, $a$ encodes the status of the
  robot arm (empty or carrying a key), $l$ encodes the status of the
  locked location (locked or open), and $k_1$ and $k_2$ encode the
  locations of the two keys.}

A worse case is shown in Fig.~\ref{figure:causal-graph-grid}, which
shows an example from the {\grid} domain with an arbitrary number of
locations, of which a single one is locked. There are two keys, one of
which can unlock the locked location. Eliminating cycles here requires
a few minor relaxations regarding the status of the robot arm (empty
or non-empty), but also one major simplification, namely the
elimination of the arc from $l$ to $r$ representing the fact that the
robot can only enter the locked location if it has been unlocked.

As a (nearly) worst-case example, consider a task in the
{\blocksworld} domain (no figure). A typical MPT encoding uses one
state variable $h$ for encoding whether or not the hand is empty and
two state variables per block in the task: For the $i$-th block, $t_i$
encodes whether or not the block is lying on the table, and $b_i$
encodes which block is lying on top of it, or if it is clear or being
held by the arm. In the causal graph of such a task, variable $h$ has
ingoing arcs from and outgoing arcs to all other state variables, and
all state variables $b_i$ are connected to each other in both
directions. Only the state variables $t_i$ have a slightly simpler
connection structure, being only connected to $h$ and to $b_i$ for the
same value of $i$. Any relaxation of the problem that eliminates
cycles from the causal graph loses a large amount of information, and
it is not surprising that the {\depot} domain, which includes a
{\blocksworld} subproblem, is the one for which the precursor of Fast
Downward fared worst \cite{malte-icaps-cg}. Still, it should be
pointed out that planners that ignore delete effects have similar
problems with \blocksworld-like domains, as the comparison between the FF
and causal graph heuristics in the same article shows.

\subsection{Successor Generators and Axiom Evaluators}
\label{section:successor-generators}

In addition to good heuristic guidance, a forward searching planning
system needs efficient methods for generating successor states if it
is to be applied to the benchmark suite from the international
planning competitions. For some domains, our causal graph heuristic or
other popular methods like the FF heuristic provide excellent goal
estimates, yet still planning can be too time-consuming because of
very long plans and vast branching factors.

The variant of best-first search implemented in Fast Downward does not
compute the heuristic estimate for each state that is generated.
Essentially, heuristic evaluations are only computed for closed nodes,
while computation is deferred for nodes on the search
frontier. For domains with strong heuristic guidance and large
branching factors, the number of nodes on the frontier can by far
dominate the number of nodes in the closed set. As a case in point,
consider the problem instance {\satellite} \#29. For solving this
task, the default configuration of Fast Downward only computes
heuristic estimates for 67\,597 world states while adding
107\,233\,381 states to the frontier. Clearly, determining the set of
applicable operators quickly is of critical importance in such a
scenario.

In some {\satellite} tasks, there are almost 1\,000\,000 ground
operators, so we should try to avoid individually checking each
operator for applicability.  Similarly, in the biggest {\psr} tasks,
more than 100\,000 axioms must be evaluated in each state to compute
the values of the derived variables, so this computation must be made
efficient. For these purposes, Fast Downward uses two data structures
called \emph{successor generators} and \emph{axiom evaluators}.

\subsubsection{Successor Generators}

Successor generators are recursive data structures very similar to
decision trees. The internal nodes have associated conditions, which
can be likened to the decisions in a decision tree, and the leaves
have associated operator lists which can be likened to a set of
classified samples in a decision tree leaf. They are formally defined
as follows.

\begin{definition}{successor-generator}{Successor generators}
  A \definedTerm{successor generator} for an MPT $\Pi = \langle
  \mathcal V, s_0, s_\star, \mathcal A, \mathcal O\rangle$ is a tree
  consisting of \definedTerm{selector nodes} and
  \definedTerm{generator nodes}.

  A selector node is an internal node of the tree. It has an
  associated variable $v \in \mathcal V$ called the
  \definedTerm{selection variable}. Moreover, it has $|\mathcal D_v|
  + 1$ children accessed via labelled edges, one edge labelled $v = d$
  for each value $d \in \mathcal D_v$, and one edge labelled $\top$.
  The latter edge is called the \definedTerm{don't care edge} of the
  selector.

  A generator node is a leaf node of the tree. It has an associated
  set of operators from $\mathcal O$ called the set of
  \definedTerm{generated operators}.

  Each operator $o \in \mathcal O$ must occur in exactly one generator
  node, and the set of edge labels leading from the root to this
  node (excluding don't care edges) must equal the precondition of $o$.
\end{definition}

Given a successor generator for an MPT $\Pi$ and a state $s$ of $\Pi$,
we can compute the set of applicable operators in $s$ by traversing
the successor generator as follows, starting from the root:
\begin{itemize}
\item At a selector node with selection variable $v$, follow the edge
  $v = s(v)$ and the don't care edge.
\item At a generator node, report the generated operators as
  applicable.
\end{itemize}

To build a successor generator for $\Pi$, we apply a
top-down algorithm which considers the task variables in an arbitrary
order $v_1 \prec v_2 \prec \dots \prec v_n$. At the root node, we
choose $v_1$ as selection variable and classify the set of operators
according to their preconditions with respect to $v_1$. Operators with
a precondition $v_1=d$ will be represented in the child of the root
accessed by the edge with the corresponding label, while operators without
preconditions on $v_1$ will be represented in the child of the root
accessed by the don't care edge. In the children of the root, we
choose $v_2$ as selection variable, in the grandchildren $v_3$, and so
on.

There is one exception to this rule to avoid creating unnecessary
selection nodes: If no operator in a certain branch of the tree has a
condition on $v_i$, then $v_i$ is not considered as a selection
variable in this branch. The construction of a branch ends when all
variables have been considered, at which stage a generator node is
created for the operators associated with that branch.

\subsubsection{Axiom Evaluators}

Axiom evaluators are a simple data structure used for efficiently
implementing the well-known \emph{marking algorithm} for propositional
Horn logic \cite{horn-marking}, extended and modified for the
layered logic programs that correspond to the axioms of an MPT.
They consist of two parts. Firstly, an indexing data structure maps a
given variable/value pairing and a given axiom layer to the set of
axioms in the given layer in whose body the pairing appears. Secondly, a
set of counters, one for each axiom, counts the number of conditions
of the axiom that have not yet been derived.

  \begin{figure}[tbp]
    \begin{center}
      \input{figures/evaluate-axioms-efficiently}
    \end{center}
    \caption{Computing the values of
  the derived variables in a given planning state.}
    \label{figure:evaluate-axioms-efficiently}
  \end{figure}

Within Fast Downward, axioms are evaluated in two steps. First, all
derived variables are set to their default value $\bot$. Second,
algorithm \emph{evaluate-axiom-layer}
(Fig.~\ref{figure:evaluate-axioms-efficiently}) is executed for each
axiom layer in sequence to determine the final values of the derived
variables.

We assume that the reader is familiar enough with the marking
algorithm not to require much explanation, so we only point out that
the test whether or not an axiom is ready to trigger is implemented by
means of a queue in which axioms are put as soon as their counter
reaches 0. The actual implementation of \emph{evaluate-axiom-layer}
within Fast Downward initializes axiom counters slightly more
efficiently than indicated by the pseudo-code. However, this is a
minor technical detail, so we turn to the remaining piece of Fast
Downward's architecture, the search component.

\section{Search}
\label{section:search}

Unlike the translation and knowledge compilation components, for which
there is only a single mode of execution, the search component of Fast
Downward can perform its work in various alternative ways. There are
three basic search algorithms to choose from:

\begin{enumerate}
\item \emph{Greedy best-first search:} This is the standard textbook
  algorithm \cite{russell-norvig}, modified with a technique called
  \emph{deferred heuristic evaluation} to mitigate the negative
  influence of wide branching. We have also extended the algorithm to
  deal with \emph{preferred operators}, similar to FF's helpful
  actions \cite{ff}. We discuss greedy best-first search in Section
  \ref{section:greedy-best-first-search}.  Fast Downward uses this
  algorithm together with the causal graph heuristic, discussed
  in Section \ref{section:cg-heuristic}.
\item \emph{Multi-heuristic best-first search:} This is a variation of
  greedy best-first search which evaluates search states using multiple
  heuristic estimators, maintaining separate open lists for each. Like
  our variant of greedy best-first search, it supports the use of
  \emph{preferred operators}. Multi-heuristic best-first search is
  discussed in Section
  \ref{section:multi-heuristic-best-first-search}. Fast Downward uses
  this algorithm together with the causal graph and FF heuristics,
  discussed in Sections \ref{section:cg-heuristic} and
  \ref{section:ff-heuristic}.
\item \emph{Focused iterative-broadening search:} This is a simple
  search algorithm that does not use heuristic estimators, and instead
  reduces the vast set of search possibilities by focusing on a
  limited operator set derived from the causal graph. It is an
  experimental algorithm; in the future, we hope to further develop
  the basic idea of this algorithm into a more robust method. Focused
  iterative-broadening search is discussed in Section
  \ref{section:focused-iterative-broadening-search}.
\end{enumerate}

For the two heuristic search algorithms, a second choice must be made
regarding the use of \emph{preferred operators}. There are five
options supported by the planner:
\begin{enumerate}
\item Do not use preferred operators.
\item Use the \emph{helpful transitions} of the causal graph heuristic
  as preferred operators.
\item Use the \emph{helpful actions} of the FF heuristic as preferred
  operators.
\item Use helpful transitions as preferred operators, falling back to
  helpful actions if there are no helpful transitions in the current
  search state.
\item Use both helpful transitions and helpful actions as preferred
  operators.
\end{enumerate}

Each of these five options can be combined with any of the two
heuristic search algorithms, so that there is a total of eleven
possible settings for the search component, ten using one of the
heuristic algorithms and one using focused iterative-broadening search.

In addition to these basic settings, the search component can be
configured to execute several alternative configurations in parallel
by making use of an internal scheduler. Both configurations of Fast
Downward that participated in IPC4 made use of this feature by running
one configuration of the heuristic search algorithms in parallel with
focused iterative-broadening search. As its heuristic search
algorithm, the configuration \emph{Fast Downward} employed greedy
best-first search with helpful transitions, falling back to helpful
actions when necessary (option 4.). The configuration \emph{Fast
Diagonally Downward} employed multi-heuristic best-first search using
helpful transitions and helpful actions as preferred operators (option
5.).

To avoid confusion between the complete Fast Downward planning system
and the particular configuration called ``Fast Downward'', we will
refer to the IPC4 planner configurations as \emph{FD} and \emph{FDD}
for the rest of this paper. The name of the planning system as a whole
is never abbreviated.

\subsection{The Causal Graph Heuristic}
\label{section:cg-heuristic}

The \emph{causal graph heuristic} is the centrepiece of Fast
Downward's heuristic search engine. It estimates the cost of reaching
the goal from a given search state by solving a number of subproblems
of the planning task which are derived by looking at small ``windows''
of the (pruned) causal graph.
For some additional intuitions about the design
of the heuristic and a discussion of theoretical aspects, we
refer to the article in which the heuristic was first introduced
\cite{malte-icaps-cg}.

\subsubsection{Conceptual View of the Causal Graph Heuristic}

For each state variable $v$ and each pair of values $d, d' \in
\mathcal D_v$, the causal graph heuristic computes a heuristic estimate
$\textit{cost}_v(d,d')$ for the cost of changing the value of $v$ from
$d$ to $d'$, assuming that all other state variables carry the same
values as in the current state. (This is a simplification. Cost
estimates are not computed for state variables $v$ or values $d$ for
which they are never required. We ignore this fact when discussing the
heuristic on the conceptual level.) The heuristic estimate of a given
state $s$ is the sum over the costs $\textit{cost}_v(s(v),
s_\star(v))$ for all variables $v$ for which a goal condition
$s_\star(v)$ is defined.

Conceptually, cost estimates are computed one variable after the
other, traversing the (pruned) causal graph in a bottom-up fashion.
By bottom-up, we mean that we start with the variables that have no
predecessors in the causal graphs; we call this order of computation
``bottom-up'' because we consider variables that can change their
state of their own accord \emph{low-level}, while variables whose
state transitions require the help of other variables have more
complex transition semantics and are thus considered
\emph{high-level}. Note that in our figures depicting causal graphs,
\emph{high-level} variables are typically displayed near the
\emph{bottom}.

For variables without predecessors in the causal graph,
$\textit{cost}_v(d,d')$ simply equals the cost of a shortest path from
$d$ to $d'$ in the (pruned) domain transition graph $\textit{DTG}(v)$.
For other variables, cost estimates are also computed by graph search
in the domain transition graph. However, the conditions of transitions
must be taken into account during path planning, so that in addition
to counting the number of transitions required to reach the
destination value, we also consider the costs for achieving the value
changes of the other variables necessary to set up the transition
conditions.

The important point here is that in computing the values
$\textit{cost}_v(d,d')$, we completely consider all interactions of
the state variable $v$ with its predecessors in the causal graph. If
changing the value from $d$ to $d'$ requires several steps and each of
these steps has an associated condition on a variable $v'$, then we
realize that $v'$ must assume the values required by those conditions
\emph{in sequence}. For example, if $v$ represents a package in a
transportation task that must be moved from $A$ to $B$ by means of a
vehicle located at $C$, then we recognize that the vehicle must first
move from $C$ to $A$ and then from $A$ to $B$ in order to drop the
package at $B$. This is very different to the way HSP- or FF-based
heuristics work on such examples. However, we only consider
interactions with the \emph{immediate} predecessors of $v$ in the
causal graph. Interactions that occur via several graph layers
are not captured by the heuristic estimator.

In essence, we compute $\textit{cost}_v(d,d')$ by solving a particular
subproblem of the MPT, induced by the variable $v$ and its
predecessors in the pruned causal graph. For this subproblem, we
assume that $v$ is initially set to $d$, we want $v$ to assume the
value $d'$, and all other state variables carry the same value as in
the current state. We call this planning problem the \emph{local
subproblem for $v$, $d$ and $d'$}, or the \emph{local subproblem for
$v$ and $d$} if we leave the target value $d'$ open.

  \begin{figure}[tbp]
    \begin{center}
      \input{figures/compute-all-costs}
    \end{center}
    \caption{The \emph{compute-costs-bottom-up}
  algorithm, a high-level description of the causal graph heuristic.}
    \label{figure:compute-all-costs}
  \end{figure}

For a formalization of these intuitive notions of how the \emph{cost}
estimates are generated, consider the pseudo-code in
Fig.~\ref{figure:compute-all-costs}. It does not reflect the way the
heuristic values are actually computed within Fast Downward; the
algorithm in the figure would be far too expensive to evaluate for
each search state. However, it computes the same cost values as Fast
Downward does, provided that the algorithm generating the plans
$\pi$ in the last line of the algorithm is the same one as the one
used for the ``real'' cost estimator.

\subsubsection{Computation of the Causal Graph Heuristic}

The actual computation of the causal graph heuristic traverses the causal graph
in a top-down direction starting from the goal variables, rather than
bottom-up starting from variables without causal predecessors. In
fact, this top-down traversal of the causal graph is the reason for
Fast Downward's name.

Computing cost estimates in a top-down traversal implies that while
the algorithm is computing plans for local subproblems of a given
variable, it typically does not yet know the costs for changing the
state of its causal predecessors. The algorithm \emph{compute-costs}
addresses this by evaluating the cost values of dependent variables
through recursive invocations of itself.

For a given variable-value pairing $v=d$, we always compute the costs
$\textit{cost}_v(d,d')$ for all values of $d' \in \mathcal D_v$ at the
same time, similar to the way Dijkstra's algorithm computes the
shortest path not from a single source to a single destination vertex,
but from a single source to all possible destination vertices.
Computing the costs for all values of $d'$ is not (much) more
expensive than computing only one of these values, and once all cost
values have been determined, we can cache them and re-use them if they
are needed again later during other parts of the computation of the
heuristic value for the current state.

In fact, the similarity to shortest path problems is not superficial
but runs quite deeply. If we ignore the recursive calls for computing
cost values of dependent variables, \emph{compute-costs} is basically
an implementation of Dijkstra's algorithm for the single-source
shortest path problem on domain transition graphs. The only difference
to the ``regular'' algorithm lies in the fact that we do not know the
cost for using an arc in advance. Transitions of derived variables
have a base cost of 0 and transitions of fluents have a base cost of
1, but in addition to the base cost, we must pay the cost for
achieving the conditions associated with a transition. However, the
cost for achieving a given condition $v'=e'$ depends on the current
value $e$ of that state variable at the time the transition is taken.
Thus, we can only compute the real cost for a transition once we know
the values of the dependent state variables in the relevant situation.

Of course, there are many different ways of taking transitions through
domain transition graphs, all potentially leading to different values for the
dependent state variables. When we first introduced the causal graph heuristic, we
showed that deciding plan existence for the local subproblems is
\NP-complete \cite{malte-icaps-cg}, so we are content with an approach
that does not lead to a complete planning algorithm, as long as it
works well for the subproblems we face in practice.

The approach we have chosen is to achieve each value of state variable
$v$ in the local subproblem for $v$ and $d$ as quickly as possible,
following a greedy policy. In the context of the Dijkstra algorithm,
this means that we start by finding the cheapest possible plan to
make a transition from $d$ to some other value $d'$. Once we have
found the cheapest possible plan $\pi_{d'}$, we commit to it,
annotating the vertex $d'$ of the domain transition graph with the
local state obtained by applying plan $\pi_{d'}$ to the current state.
In the next step, we look for the cheapest possible plan to achieve
another value $d''$, by either considering transitions that start from
the initial value $d$, or by considering transitions that continue the
plan $\pi_{d'}$ by moving to a neighbour of $d'$. This process is
iterated until all vertices of the domain transition graph have been
reached or no further progress is possible.

  \begin{figure}[tbp]
    \begin{center}
      \input{figures/compute-costs}
    \end{center}
    \caption{Fast Downward's implementation of the causal graph
  heuristic: the \emph{compute-costs} algorithm for computing the
  estimates $cost_v(d, d')$ for all values $d' \in \mathcal D_v$ in a
  state $s$ of an MPT $\Pi$.}
    \label{figure:compute-costs}
  \end{figure}

Our implementation follows Dijkstra's algorithm
(Fig.~\ref{figure:compute-costs}). We have implemented the priority
queue as a vector of buckets for maximal speed and use a cache to
avoid generating the same $\textit{cost}_v(d,d')$ value twice for the
same state. In addition to this, we use a global cache that is shared
throughout the whole planning process so that we need to compute the
values $\textit{cost}_v(d,d')$ for variables $v$ with few ancestors in
the pruned causal graph only once. (Note that $\textit{cost}_v(d,d')$
only depends on the current values of the ancestors of $v$.)

Apart from these and some other technical considerations,
Fig.~\ref{figure:compute-costs} gives an accurate account of Fast
Downward's implementation of the causal graph heuristic. For more details,
including complexity considerations and a worked-out example, we refer
to the original description of the algorithm \cite{malte-icaps-cg}.

\subsubsection{States with Infinite Heuristic Value}

We noted that Fast Downward uses an incomplete planning algorithm
for determining solutions to local planning problems. Therefore, there
can be states $s$ with $\textit{cost}_v(s(v), s_\star(v)) =
\infty$ even though the goal condition $v=s_\star(v)$ can still be
reached. This means that we cannot trust infinite values returned by
the causal graph heuristic. In our experience, states with infinite heuristic
evaluation from which it is still possible to reach the goal are rare,
so we indeed treat such states as \emph{dead ends}.

If it turns out that \emph{all} states at the search frontier are dead
ends, we cannot make further progress with the causal graph heuristic.
In this case, we use a sound dead-end detection routine to verify the
heuristic assessment. If it turns out that all frontier states are
indeed dead ends, then we report the problem as unsolvable. Otherwise,
search is restarted with the FF heuristic (cf.~Section
\ref{section:ff-heuristic}), which is sound for purposes of dead-end
detection.\footnote{In practice, we have never observed the causal
graph heuristic to fail on a solvable task. Therefore, the fallback
mechanism is only used for some unsolvable tasks in the {\miconicfull}
domain which are not recognized by our dead-end detection technique.}

The dead-end detection routine has been originally developed for
STRIPS-like tasks. However, extending it to full MPTs is easy; in
fact, no changes to the core algorithm are required, as it works at
the level of domain transition graphs and is still sound when applied
to tasks with conditional effects and axioms. Since it is not a
central aspect of Fast Downward, we do not discuss it here, referring
to our earlier work instead \cite{malte-icaps-cg}.

\subsubsection{Helpful Transitions}

Inspired by Hoffmann's very successful use of \emph{helpful actions}
within the FF planner \cite{ff}, we have extended our algorithm for
computing the causal graph heuristic so that in addition to the heuristic
estimate, it also generates a set of applicable operators considered
useful for steering search towards the goal.

To compute helpful actions in FF, Hoffmann's algorithm generates a
plan for the relaxed planning task defined by the current search state
and considers those operators \emph{helpful} which belong to the
relaxed plan and are applicable in the current state.

Our approach follows a similar idea. After computing the heuristic
estimate $\textit{cost}_v(s(v), s_\star(v))$ for a variable $v$ for
which a goal condition is defined, we look into the domain transition
graph of $v$ to trace the path of transitions leading from $s(v)$ to
$s_\star(v)$ that gave rise to the cost estimate. In particular, we
consider the first transition on this path, starting at $s(v)$. If
this transition corresponds to an applicable operator, we consider
that operator a \emph{helpful transition} and continue to check the
next goal. If the transition does not correspond to an applicable
operator because it has associated conditions of the form $v'=e'$
which are not currently satisfied, then we recursively look for
helpful transitions in the domain transition graph of each such
variable $v'$, checking the path that was generated during the
computation of $\textit{cost}_{v'}(s(v'), e')$.

The recursive process continues until we have found all helpful
transitions. Unlike the case for FF, where helpful actions can be
found for all non-goal states, we might not find any helpful
transition at all. It may be the case that a transition does not
correspond to an applicable operator even though it has no associated
conditions; this can happen when some operator preconditions are not
represented in the pruned domain transition graph due to cycles in the
causal graph. Even so, we have found helpful transitions to be a
useful tool in guiding our best-first search algorithms.

\subsection{The FF Heuristic}
\label{section:ff-heuristic}

The \emph{FF heuristic} is named after Hoffmann's planning algorithm
of the same name, in the context of which it was originally introduced
\cite{ff}. It is based on the notion of \emph{relaxed planning tasks}
that ignore \emph{negative interactions}. In the context of MPTs,
ignoring negative interactions means that we assume that each state
variable can hold several values simultaneously. An operator effect or
axiom that sets a variable $v$ to a value $d$ in the original task
corresponds to an effect or axiom that \emph{adds} the value $d$ to
the range of values assumed by $v$ in the relaxed task. A condition
$v=d$ in the original task corresponds to a condition requiring $d$ to
be an element of the set of values currently assumed by $v$ in the
relaxed task.

It is easy to see that applying some operator in a solvable relaxed
planning task can never render it unsolvable. It can only lead to more
operators being applicable and more goals being true, if it has any
significant effect at all. For this reason, relaxed planning tasks can
be solved efficiently, even though optimal solutions are still
\NP-hard to compute \cite{complexity-bylander}. A plan for the
relaxation of a planning task is called a \emph{relaxed plan} for
that task.

The FF heuristic estimates the goal distance of a world state by
generating a relaxed plan for the task of reaching the goal from this
world state. The number of operators in the generated plan is then used as the
heuristic estimate. Our implementation of the FF heuristic does not
necessarily generate the same, or even an equally long, relaxed plan
as FF. In our experiments, this did not turn out to be problematic, as
both implementations appear to be equally informative.

While the FF heuristic was originally introduced for ADL domains,
extending it to tasks involving derived predicates is
straight-forward. One possible extension is to simply assume that each
derived predicate is initially set to its default value $\bot$ and
treat axioms as relaxed operators of cost 0. In a slightly more
complicated, but also more accurate approach, derived variables are
initialized to their actual value in a given world state, allowing the
relaxed planner to achieve the value $\bot$ (or other values) by
applying the transitions of the extended domain transition graph of
the derived variable. We have followed the second approach.

In addition to heuristic estimates, the FF heuristic can also be
exploited for restricting or biasing the choice of operators to apply
in a given world state $s$. The set of \emph{helpful actions} of $s$
consists of all those operators of the relaxed plan computed for $s$
that are applicable in that state. As mentioned in the introduction to
this section, Fast Downward can be configured to treat helpful actions
as preferred operators.

There is a wealth of work on the FF heuristic in the literature, so we
do not discuss it further. For a more thorough treatment, we 
point to the references
\cite{ff,joerg-topology,hoffmann:aips-02,hoffmann:jair2005}.

\subsection{Greedy Best-First Search in Fast Downward}
\label{section:greedy-best-first-search}

Fast Downward uses \emph{greedy best-first search with a closed list}
as its default search algorithm. We assume that the reader is familiar
with the algorithm and refer to the literature for details
\cite{russell-norvig}.

Our implementation of greedy best-first search differs from the
textbook algorithm in two ways. First, it can treat helpful
transitions computed by the causal graph heuristic or helpful actions
computed by the FF heuristic as \emph{preferred operators}. Second, it
performs \emph{deferred heuristic evaluation} to reduce the influence
of large branching factors. We now turn to describing these two search
enhancements.

\subsubsection{Preferred Operators}

To make use of helpful transitions computed by the causal graph heuristic
or helpful actions computed by the FF heuristic, our variant of
greedy best-first search supports the use of so-called \emph{preferred
operators}.
The set of preferred operators of a given state is a subset of the set
of applicable operators for this state. Which operators are considered
preferred depends on the settings for the search component, as
discussed earlier. The intuition behind preferred operators is that a
randomly picked successor state is more likely to be closer to the
goal if it is generated by a preferred operator, in which case we call
it a \emph{preferred successor}. Preferred successors should be
considered before non-preferred ones on average.

Our search algorithm implements this preference by maintaining two
separate open lists, one containing \emph{all} successors of expanded
states and one containing \emph{preferred} successors exclusively. The search
algorithm alternates between expanding a regular successor and a
preferred successor. On even iterations it will consider the one open
list, on odd iterations the other. No matter which open list a state
is taken from, all its successors are placed in the first open list,
and the preferred successors are additionally placed in the second
open list.  (Of course we could limit the first open list to only
contain non-preferred successors; however, typically the total number
of successors is vast and the number of preferred successors is tiny.
Therefore, it is cheaper to add all successors to the first open list
and detect duplicates upon expansion than scan through the list of
successors determining for each element whether or not it is
preferred.)

Since the number of preferred successors is smaller than the total
number of successors, this means that preferred successors are
typically expanded much earlier than others. This is especially
important in domains where heuristic guidance is weak and a lot of
time is spent exploring plateaus. When faced with plateaus, Fast
Downward's open lists operate in a first-in-first-out fashion. (In
other words: For a constant heuristic function, our search algorithm
behaves like breadth-first search.) Preferred operators typically
offer much better chances of escaping from plateaus since they lead
to significantly lower effective branching factors.

\subsubsection{Deferred Heuristic Evaluation}

Upon expanding a state $s$, the textbook version of greedy best-first
search computes the heuristic evaluation of all successor states of
$s$ and sorts them into the open list accordingly. This can be
wasteful if $s$ has many successors and heuristic evaluations are
costly, two conditions that are often true for heuristic search
approaches to planning.

This is where our second modification comes into play. If a successor
with a better heuristic estimate than $s$ is generated early and leads
to a promising path towards the goal, we would like to avoid
generating the other successors. Let us assume that $s$ has 1000
successors, and that $s'$, the 10th successor of $s$ being generated,
has a better heuristic estimate than $s$. Furthermore, let us assume
that the goal can be reached from $s'$ on a path with non-increasing
heuristic estimates. Then we would like to avoid computing heuristic
values for the 990 later successors of $s$ altogether.

Deferred heuristic evaluation achieves this by \emph{not} computing
heuristic estimates for the successors of an expanded state $s$
immediately. Instead, the successors of $s$ are placed in the open
list together with the heuristic estimate \emph{of state $s$}, and
their own heuristic estimates are only computed when and if they are
expanded, at which time it is used for sorting \emph{their} successors
into the open list, and so on. In general, each state is sorted into
the open list according to the heuristic evaluation of its parent,
with the initial state being an exception. In fact, we do not need to
put the successor state itself into the open list, since we do not
require its representation before we want to evaluate its heuristic
estimate. Instead, we save memory by storing only a reference to the
parent state and the operator transforming the parent state into the
successor state in the open list.

It might not be clear how this approach can lead to significant
savings in time, since deferred evaluation also means that information
is only available later. The potential savings become most apparent
when considering deferred heuristic evaluation together with the use
of preferred operators: If an improving successor $s'$ of a state $s$
is reached by a preferred operator, it is likely that it will be
expanded (via the second open list) long before most other successors
--- or even most siblings --- of $s$. In the situation described
above, where there exists a non-increasing path from $s'$ to the goal,
heuristic evaluations will never be computed for most successors of
$s$. In fact, deferred heuristic evaluation can significantly improve
search performance even when preferred operators are not used,
especially in tasks where branching factors are large and the
heuristic estimate is informative.

At first glance, deferred heuristic evaluation might appear related to
another technique for reducing the effort of expanding a node within a
best-first search algorithm, namely A${}^*$ with Partial Expansion
\cite{astar-partial-expansion}. However, this algorithm is designed
for reducing the \emph{space} requirements of best-first search at the
expense of additional heuristic evaluations: When expanding a node,
A${}^*$ with Partial Expansion computes the heuristic value of
\emph{all} successors, but only stores those in the open queue whose
heuristic values fall below a certain \emph{relevance threshold}. In
later iterations, it might turn out that the threshold was chosen too
low, in which case the node needs to be re-expanded and the heuristic
values of its successors re-evaluated. In general, A${}^*$ with
Partial Expansion will never compute fewer heuristic estimates than
standard A${}^*$, but it will usually require less memory.

However, for heuristic search approaches to planning (and certainly
for Fast Downward), heuristic evaluations are usually so costly in
time that memory for storing open and closed lists is not a limiting
factor. We are thus willing to trade off memory with time in the
opposite way: Deferred heuristic evaluation normally leads to more
node expansions and higher space requirements than standard best-first
search because the heuristic values used for guiding the search are
less informative (they evaluate the predecessor of a search node
rather than the node itself). However, heuristic computations are only
required for nodes that are actually removed from the open queue
rather than for all nodes on the fringe, and the latter are usually
significantly more numerous.

\subsection{Multi-Heuristic Best-First Search}
\label{section:multi-heuristic-best-first-search}

As an alternative to greedy best-first search, Fast Downward supports
an extended algorithm called \emph{multi-heuristic best-first search}.
This algorithm differs from greedy best-first search in its use of
multiple heuristic estimators, based on our observation that different
heuristic estimators have different weaknesses. It may be the case
that a given heuristic is sufficient for directing the search towards
the goal except for one part of the plan, where it gets stuck on a
plateau. Another heuristic might have similar characteristics, but get
stuck in another part of the search space.

Various ways of combining heuristics have been proposed in the
literature, typically adding together or taking the maximum of the
individual heuristic estimates. We believe that it is often beneficial
\emph{not} to combine the different heuristic estimates into a single
numerical value. Instead, we propose maintaining a \emph{separate} open list
for each heuristic estimator, which is sorted according to the
respective heuristic. The search algorithm alternates between
expanding a state from each open list. Whenever a state is expanded,
estimates are calculated according to \emph{each} heuristic, and the
successors are put into each open list.

When Fast Downward is configured to use multi-heuristic best-first
search, it computes estimates both for the causal graph heuristic and FF
heuristic, maintaining two open lists. Of course, the approach can be
combined with the use of preferred operators; in this case, the search
algorithm maintains four open lists, as each heuristic distinguishes
between normal and preferred successors.

\subsection{Focused Iterative-Broadening Search}
\label{section:focused-iterative-broadening-search}

The \emph{focused iterative-broadening search} algorithm is the most
experimental piece of Fast Downward's search arsenal. In its present
form, the algorithm is unsuitable for many planning domains,
especially those containing comparatively few different goals. Yet we
think that it might contain the nucleus for a successful approach to
domain-independent planning which is very different to most current
methods, so we include it for completeness and as a source of
inspiration.

The algorithm is intended as a first step towards developing search
techniques that emphasize the idea of using heuristic criteria
locally, for limiting the set of operators to apply, rather than
globally, for choosing which states to expand from a global set of
open states. We made first experiments in this direction after
observing the large boost in performance that can be obtained by using
preferred operators in heuristic search. The algorithm performed
surprisingly well in some of the standard benchmark domains, while
performing badly in most others.

As the name suggests, the algorithm \emph{focuses} the search by
concentrating on one goal at a time, and by restricting its attention
to operators which are supposedly important for reaching that goal:

\begin{definition}{distance}{Modification distances}
Let $\Pi$ be an MPT, let $o$ be an operator of $\Pi$, and let $v$ be a
variable of $\Pi$.

The \definedTerm{modification distance} of $o$ with respect to $v$ is
defined as the minimum, over all variables $v'$ that occur as affected
variables in the effect list of $o$, of the distance from $v'$ to $v$ in
$\textit{CG}(\Pi)$.
\end{definition}

For example, operators that modify $v$ directly have a modification
distance of 0 with respect to $v$, operators that modify variables
which occur in preconditions of operators modifying $v$ have a
modification distance of 1, and so on. We assume that in order to
change the value of a variable, operators with a low modification
distance with respect to this variable are most useful.

  \begin{figure}[tbp]
    \begin{center}
      \input{figures/reach-one-goal}
    \end{center}
    \caption{The \emph{reach-one-goal}
  procedure for reaching a state with $v=d$. The value
  \emph{max-threshold} is equal to the maximal modification distance
  of any operator with respect to $v$.}
    \label{figure:reach-one-goal}
  \end{figure}

Fig.~\ref{figure:reach-one-goal} shows the \emph{reach-one-goal}
procedure for achieving a single goal of an MPT. For the time being,
assume that the $\cond$ parameter is always $\emptyset$. The procedure
makes use of the assumption that high modification distance implies
low usefulness in two ways. First, operators with high modification
distance with respect to the goal variable are considered to have a
higher associated cost, and are hence applied less frequently. Second,
operators whose modification distance is beyond a certain threshold
are forbidden completely. Instead of choosing a threshold a priori,
the algorithm first tries to find a solution with the lowest possible
threshold of 0, increasing the threshold by 1 whenever the previous
search has failed. The \emph{uniform-cost-search} algorithm mentioned
in Fig.~\ref{figure:reach-one-goal} is the standard textbook method
\cite{russell-norvig}.

Although we were ignorant of this fact at the time our algorithm was
conceived, the core idea of \emph{reach-one-goal} is not new: Ginsberg
and Harvey \citeyear{ginsberg-harvey:aij92} present a search technique
called \emph{iterative broadening}, which is also based on the idea of
repeatedly doing a sequence of uninformed searches with an
ever-growing set of operators. Their work demonstrates the superiority
of iterative broadening over standard depth-bounded search both
empirically and analytically under the reasonable assumption that the
choices made at each branching point are equally
important.\footnote{See the original analysis for a precise definition
  of ``equally important'' \cite{ginsberg-harvey:aij92}. While
  Ginsberg and Harvey's assumption is certainly not valid in practice,
  we find it much more convincing than the competing model where goal
  states are uniformly distributed across the search fringe.} The
original iterative broadening algorithm applies to scenarios without
any knowledge of the problem domain, so it chooses the set of
operators which may be applied at every search node randomly, rather
than using heuristic information from the causal graph as in our case.
However, Ginsberg and Harvey already discuss the potential
incorporation of heuristics into the operator selection. The
introduction of operator costs (in the form of modification distances)
is new, but it is a fairly straightforward extension where heuristic
information is available.

The focused iterative-broadening search algorithm is based on the
\emph{reach-one-goal} method; the idea is to achieve the goals of the
planning task one after the other, by using the \emph{reach-one-goal}
algorithm as the core subroutine for satisfying individual goals.
Since it is not obvious what a good order of achieving the goals would
be, one invocation of \emph{reach-one-goal} is started for each goal
in parallel. As each one-goal solver focuses on the (supposedly)
relevant operators for reaching its particular goal, there is hope
that the number of states considered before a goal is reached is
small. Once one of the one-goal solvers reaches its goal, the
resulting plan is reported and all sub-searches are stopped. The
overall search algorithm commits to this part of the plan; the
situation in which the first goal has been reached is considered a new
initial state.

From this situation, we try to satisfy the second goal, by once more
starting parallel invocations of \emph{reach-one-goal} for
each possible second goal. Of course, this can lead to a situation
where the search algorithm oscillates between goals, first achieving 
goal $a$, then abandoning it in favour of goal $b$, without any sign of
making real progress. Therefore, we demand that \emph{reach-one-goal}
achieves the second goal \emph{in addition} to the one we reached
first, by setting the $\cond$ argument accordingly. Once two goals
have been reached, the sub-searches are again stopped,
sub-searches for the third goal are started, and so on, until all
goals have been reached.

In some sense, our focusing technique is similar to the beam search
algorithm \cite{beam-search}, which also performs a fixed number of
concurrent searches to avoid committing to a particular path in the
search space too early. Beam search uses a heuristic function to
evaluate which branches of search should be abandoned and where new
branches should be spawned. While focused iterative-broadening search
does not appear to use heuristic evaluations at first glance, the
number of satisfied goals of a state is used as an evaluation
criterion in essentially the same way. One important difference to
beam search is our use of modification distances relative to a
particular goal, which means that the different ``beams'' explore the
state space in qualitatively different ways.

There is one final twist: To motivate \emph{reach-one-goal} not to
needlessly wander away from satisfied goals, we forbid applying
operators that undo any of the previously achieved goals in $\cond$.
This is an old idea called \emph{goal protection}
\cite{joslin-roach:aij89}. It is well-known that protecting goals
renders a search algorithm incomplete, even in state spaces where all
operators are reversible and local search approaches like focused
iterative-broadening search would be otherwise complete. In
particular, search must fail in planning tasks which are not
\emph{serializable} \cite{korf:aij-1987}. Therefore, if the first
solution attempt fails, the algorithm is restarted without
goal protection. The complete procedure is shown in
Fig.~\ref{figure:reach-one-goal-complete}, which concludes our
discussion of Fast Downward's search component.

  \begin{figure}[tbp]
    \begin{center}
      \input{figures/reach-one-goal-complete}
    \end{center}
    \caption{The \emph{reach-one-goal}
  procedure for reaching a state with $v=d$ (corrected).}
    \label{figure:reach-one-goal-complete}
  \end{figure}

\section{Experiments}
\label{section:experiments}

To evaluate the performance of Fast Downward, and specifically the
differences between the various configurations of the search
component, we have performed a number of experiments on the set of
benchmarks from the previous international planning competitions.  The
purpose of these experiments is to compare Fast Downward to the state
of the art in PDDL planning, and to contrast the performance of the
different search algorithms of Fast Downward (greedy best-first search
with and without preferred operators, multi-heuristic best-first
search with and without preferred operators, and focused
iterative-broadening search).

To clearly state the purpose of our experiments, let us also point out
two areas worthy of study that we do \emph{not} choose to investigate
here:
\begin{itemize}
\item We do not compare the causal graph heuristic to other
  heuristics, such as the FF or HSP heuristics. Such a comparison
  would require evaluating the different heuristics within otherwise
  identical planning systems. We have performed such an experiment
  before \cite{malte-icaps-cg} and thus prefer to dedicate this
  section to an evaluation of the complete Fast Downward planning
  system, rather than just its heuristic function.
\item We do not give a final answer to the question \emph{why} Fast
  Downward performs well or badly in the domains we analyse. Where we
  do observe bad performance, we try to give a plausible explanation
  for this, but we do not conduct a full-blown study of heuristic
  quality in the spirit of Hoffmann's work on the FF and $h^+$
  heuristics \cite{hoffmann:jair2005}. While we do believe that much
  could be learned from such an investigation, it is a major
  undertaking that would go beyond the scope of this article.
\end{itemize}

Our aim in this section is to evaluate the Fast Downward planner as a
whole, so there are a number of algorithmic questions which we do not
address. For example, one might wonder what (if any) speed-up can be
obtained by using successor generators over simpler methods
which test each operator for applicability whenever a node is
expanded. Another question concerns the extent to which deferred
heuristic evaluation affects search performance. To keep this section
at a reasonable length, we do not discuss either of these questions
here. However, we have conducted experiments addressing them, and
include their results in an electronic appendix to this
paper.\footnote{See \texttt{http://www.jair.org/}. The short summary
  is that successor generators speed up search by up to two orders of
  magnitude in extreme cases like the largest {\satellite} tasks, but
  have little impact on performance most of the time. Deferred heuristic
  evaluation is very beneficial in some domains, with speed-ups of more
  than one order of magnitude being common, is somewhat beneficial in
  the majority of domains, with speed-ups between 2 and 4, and is very
  rarely detrimental to performance.}

\subsection{Benchmark Set}

The benchmark set we use consists of all propositional planning tasks
from the fully automated tracks of the first four international
planning competitions hosted at AIPS 1998, AIPS 2000, AIPS 2002 and
ICAPS 2004. The set of benchmark domains is shown in
Fig.~\ref{figure:domains}. Altogether, the benchmark suite comprises
1442 tasks. (The numbers in Fig.~\ref{figure:domains} add up to 1462,
but the 20 {\satellite} instances that were introduced for IPC3 were
also part of the benchmark set of IPC4, so we only count them once.)

  \begin{figure}[tbp]
    \begin{center}
    \newcommand{\heading}[1]{%
      \multicolumn{1}{c}{\textbf{#1}}}
    \begin{tabular}{llcr}
    \heading{Competition}
    & \heading{Domain}
    & \heading{Class}
    & \heading{Number of tasks} \\[6pt]
    IPC1 (AIPS 1998) & \assembly & ADL & 30 \\
    & \grid & STRIPS & 5 \\
    & \gripper & STRIPS & 20 \\
    & \logistics & STRIPS & 35 \\
    & \movie & STRIPS & 30 \\
    & \mystery & STRIPS & 30 \\
    & \mysteryprime & STRIPS & 35 \\[6pt]
    IPC2 (AIPS 2000) & \blocksworld & STRIPS & 35 \\
    & \freecell & STRIPS & 60 \\
    & \logistics & STRIPS & 28 \\
    & \miconicstrips & STRIPS & 150 \\
    & \miconicsimple & ADL & 150 \\
    & \miconicfull & ADL & 150 \\
    & \schedule & ADL & 150 \\[6pt]
    IPC3 (AIPS 2002) & \depot & STRIPS & 22 \\
    & \driverlog & STRIPS & 20 \\
    & \freecell & STRIPS & 20 \\
    & \rovers & STRIPS & 20 \\
    & \satellite & STRIPS & 20 \\
    & \zenotravel & STRIPS & 20 \\[6pt]
    IPC4 (ICAPS 2004) & \airport & STRIPS & 50 \\
    & \opticaltelegraph & PDDL2.2 & 48 \\
    & \philosophers & PDDL2.2 & 48 \\
    & \pipesworldnotankage & STRIPS & 50 \\
    & \pipesworldtankage & STRIPS & 50 \\
    & \psrsmall & STRIPS & 50 \\
    & \psrmiddle & PDDL2.2 & 50 \\
    & \psrlarge & PDDL2.2 & 50 \\
    & \satellite & STRIPS & 36
    \end{tabular}

    \end{center}
    \caption{Planning domains of the first four
  international planning competitions.}
    \label{figure:domains}
  \end{figure}

We distinguish between three classes of domains:
\begin{itemize}
\item \emph{STRIPS domains:} These domains do not feature derived
  predicates or conditional effects, and all conditions appearing in
  goal and operators are conjunctions of positive literals.
\item \emph{ADL domains:} These domains make use of conditional
  effects in their operator and/or contain more general conditions
  than simple conjunctions in their goals and operators. However, they
  do not require axioms.
\item \emph{PDDL2.2 domains:} These domains use the full range of
  propositional PDDL2.2, including those features present in ADL
  domains and axioms.
\end{itemize}

At IPC4, some domains were presented in different
\emph{formulations}, meaning that the same real-world task was encoded
in several different ways. Participants were asked to only work on one
formulation per domain, being able to choose their preferred
formulation for a given domain freely. For example, the {\airport}
domain was available in a STRIPS formulation and an ADL formulation.

However, the organizers did not strictly follow the rule of
considering different encodings of the same real-world task different
\emph{formulations}, rather than different domains proper. Namely, for
the {\psrmiddle} and {\promela} domains, encodings with and without
axioms were available, and these were considered as different domains
on the grounds that the encodings without axioms were much larger and
hence likely more difficult to solve. We apply the formulation
vs.~encoding view more strictly and thus only consider one
{\psrmiddle} domain and one domain for each of the two {\promela}
variants, {\philosophers} and {\opticaltelegraph}.

Of the IPC1 benchmark set, all tasks are solvable except for 11
{\mystery} instances. Of the IPC2 benchmark set, all tasks are
solvable except for 11 {\miconicfull} instances. All IPC3 benchmarks are
solvable. For IPC4, we have not checked all instances of the
{\pipesworldtankage} domain, but we assume that all are tasks are
solvable.

If run in any of the heuristic search modes, Fast Downward proves the
unsolvability of the unsolvable {\mystery} and {\miconicfull} tasks by
using the dead-end detection routine described in our earlier article
on the causal graph heuristic \cite{malte-icaps-cg}, or in some cases
in the {\miconicfull} domain by exhaustively searching all states with
a finite FF heuristic. Of course, if an unsolvable task is proved
unsolvable by the planner, we report this as a ``successfully solved''
instance in the experimental results.

\subsection{Experimental Setup}

As discussed in Section \ref{section:search}, there are eleven
possible configurations of Fast Downward's search component. However,
not all of them are equally reasonable. For example, if we use FF's
helpful actions, it would seem wasteful not to use the FF heuristic
estimate, since these two are calculated together. Therefore, for the
greedy best-first search setup, we exclude configurations where
FF helpful actions are always computed. For the multi-heuristic
best-first search setup, we exclude configurations where only one type
of preferred operators is considered, but not the other, since this
would seem to be a very arbitrary choice. This leaves us with six
different configurations of the planner:
\begin{enumerate}
\item \textbf{G}: Use greedy best-first search without preferred
  operators.
\item \textbf{G + P}: Use greedy best-first search with helpful
  transitions as preferred operators.
\item \textbf{G + P${}^+$}: Use greedy best-first search with helpful
  transitions as preferred operators. Use helpful actions as preferred
  operators in states with no helpful transitions.
\item \textbf{M}: Use multi-heuristic best-first search without 
  preferred operators.
\item \textbf{M + P}: Use multi-heuristic best-first search with
  helpful transitions and helpful actions as preferred operators.
\item \textbf{F}: Use focused iterative-broadening search.
\end{enumerate}

We apply each of these planner configurations to each of the 1442
benchmark tasks, using a computer with a 3.066 GHz Intel Xeon CPU ---
the same machine that was used at IPC4 --- and set a memory limit of 1
GB and a timeout of 300 seconds.

To compare Fast Downward to the state of the art, we try to solve
each benchmark with the best-performing planners from the literature.
Unfortunately, this involves some intricacies: some planners are not
publicly available, and others only cover a restricted subset of PDDL2.2.
For the main experiment, we thus partition the benchmark domains into
three sets depending on which planners are available for comparison.

\subsection{Translation and Knowledge Compilation vs.~Search}

Of course, the results we report for Fast Downward include the time
spent in all three components of the planner: translation, knowledge
compilation, and search. Therefore, in the following presentation of
results, we only consider a task solved if the \emph{total} processing
time is below 300 seconds. However, we have also investigated which
tasks can be solved with a timeout of 300 seconds for the
\emph{search} component alone, allowing the other components to use an
arbitrary amount of resources. It turns out that this only makes a
difference in five cases, most of which could have been solved in a
total time below 310 seconds (Fig.~\ref{figure:total-time-above-300}).
Only in one of these five cases, a {\satellite} instance of exorbitant
size, did search take less time than the other two phases combined.
These results show that the search component is the only time-critical
part of Fast Downward in practice. Therefore, we do not report
separate performance results for the individual components.

  \begin{figure}[tbp]
    \begin{center}
    \newcommand{\heading}[1]{%
      \multicolumn{1}{c}{\textbf{#1}}}
    \begin{tabular}{@{}lllrr@{}}
    \heading{Domain}
    & \heading{Task}
    & \heading{Configuration}
    & \heading{Preprocessing}
    & \heading{Search} \\[3pt]
    {\freecell} (IPC2)  & \texttt{probfreecell-10-1} &
      \textbf{M + P}    &   9.30\,s & 298.64\,s \\
    {\grid}             & \texttt{prob05} &
      \textbf{M}        &  10.04\,s & 291.01\,s \\
    {\mysteryprime}     & \texttt{prob14} & 
      \textbf{M}        &  22.38\,s & 291.67\,s \\
    {\psrlarge}         & \texttt{p30-s179-n30-l3-f30} &
      \textbf{G + P}    &  43.43\,s & 265.29\,s \\
    {\satellite} (IPC4) & \texttt{p33-HC-pfile13} &
      \textbf{M + P}    & 180.74\,s & 169.09\,s
    \end{tabular}

    \end{center}
    \caption{Tasks which could be solved by
  some configuration of Fast Downward with a search timeout of 300
  seconds, but not with a total processing timeout of 300 seconds. The
  column ``preprocessing'' shows the total time for translation and
  knowledge compilation.}
    \label{figure:total-time-above-300}
  \end{figure}

\subsection{STRIPS Domains from IPC1--3}

Let us now present the results of the main experiment. We abstain from
listing runtimes for individual planning tasks due to the
prohibitively large amount of data. These are available as an
electronic appendix to this
article.\footnote{\texttt{http://www.jair.org/}} Instead, we report
the following information:
\begin{itemize}
  \item Tables showing the number of tasks \emph{not solved} by each
    planner within the 300 second timeout. Here, we present individual
    results for each domain.
  \item Graphs showing the number of tasks solved in a given time by
    each planner. Here, we do not present separate results for each
    domain, as this would require too many graphs.
\end{itemize}

We do not discuss plan lengths; our observations in this regard are
similar to those made for the original implementation of the causal
graph heuristic \cite{malte-icaps-cg}.

\psfrag{diagonallydownward}[r][r]{\tiny \textsf{\textbf{FDD} (Fast Downward)}}
\psfrag{downward}[r][r]{\tiny \textsf{\textbf{FD} (Fast Downward)}}
\psfrag{yahsp}[r][r]{\tiny \textsf{YAHSP}}
\psfrag{macro-ff}[r][r]{\tiny \textsf{Macro-FF}}
\psfrag{sgplan}[r][r]{\tiny \textsf{SGPlan}}
\psfrag{lpg-td}[r][r]{\tiny \textsf{LPG-TD}}

\psfrag{CG}[r][r]{\tiny \textsf{CG}}
\psfrag{FF}[r][r]{\tiny \textsf{FF}}
\psfrag{LPG}[r][r]{\tiny \textsf{LPG}}
\psfrag{Any}[r][r]{\tiny \textsf{\textbf{Any} (Fast Downward)}}
\psfrag{G+P+}[r][r]{\tiny \textsf{\textbf{G + P${}^+$} (Fast Downward)}}
\psfrag{G+P}[r][r]{\tiny \textsf{\textbf{G + P} (Fast Downward)}}
\psfrag{G}[r][r]{\tiny \textsf{\textbf{G} (Fast Downward)}}
\psfrag{M+P}[r][r]{\tiny \textsf{\textbf{M + P} (Fast Downward)}}
\psfrag{M}[r][r]{\tiny \textsf{\textbf{M} (Fast Downward)}}
\psfrag{F}[r][r]{\tiny \textsf{\textbf{F} (Fast Downward)}}

  \begin{figure}[htbp]
    \begin{center}
    \newcommand{\heading}[1]{%
      \multicolumn{1}{c}{\makebox[0.6cm]{\textbf{#1}}}}
    \newcommand{\headingrule}[1]{%
      \multicolumn{1}{c|}{\makebox[0.6cm]{\textbf{#1}}}}
    \begin{tabular}{lr|cccccc|c|ccc}
      \heading{Domain}
    & \multicolumn{1}{c|}{\textbf{\#Tasks}}
    & \heading{G}
    & \heading{G+P}
    & \heading{G+P${}^+$}
    & \heading{M}
    & \heading{M+P}
    & \headingrule{F}
    & \headingrule{Any}
    & \heading{CG}
    & \heading{FF}
    & \heading{LPG} \\[6pt]
    \blocksworld & 35
      & 0 & 0 & 0 & 0 & 0 & 17 & 0 & 0 & 4 & 0 \\
    \depot & 22
      & 12 & 13 & 13 & 12 & 8 & 11 & 7 & 14 & 3 & 0 \\
    \driverlog & 20
      & 2 & 0 & 0 & 1 & 0 & 1 & 0 & 3 & 5 & 0 \\
    {\freecell} (IPC2) & 60
      & 4 & 4 & 12 & 11 & 12 & 40 & 3 & 2 & 3 & 55 \\
    {\freecell} (IPC3) & 20
      & 0 & 0 & 5 & 1 & 2 & 14 & 0 & 0 & 2 & 19 \\
    \grid & 5
      & 1 & 2 & 1 & 1 & 0 & 4 & 0 & 1 & 0 & 1 \\
    \gripper & 20
      & 0 & 0 & 0 & 0 & 0 & 0 & 0 & 0 & 0 & 0 \\
    {\logistics} (IPC1) & 35
      & 1 & 0 & 0 & 4 & 0 & 26 & 0 & 0 & 0 & 4 \\
    {\logistics} (IPC2) & 28
      & 0 & 0 & 0 & 0 & 0 & 0 & 0 & 0 & 0 & 0 \\
    \miconicstrips & 150
      & 0 & 0 & 0 & 0 & 0 & 0 & 0 & 0 & 0 & 0 \\
    \movie & 30
      & 0 & 0 & 0 & 0 & 0 & 0 & 0 & 0 & 0 & 0 \\
    \mystery & 30
      & 1 & 2 & 1 & 0 & 0 & 13 & 0 & 1 & 12 & 15 \\
    \mysteryprime & 35
      & 0 & 0 & 0 & 2 & 0 & 14 & 0 & 1 & 3 & 7 \\
    \rovers & 20
      & 2 & 0 & 0 & 0 & 0 & 2 & 0 & 3 & 0 & 0 \\
    {\satellite} (IPC3) & 20
      & 1 & 0 & 0 & 0 & 0 & 6 & 0 & 0 & 0 & 0 \\
    \zenotravel & 20
      & 0 & 0 & 0 & 0 & 0 & 0 & 0 & 0 & 0 & 0 \\[6pt]
    \textbf{Total} & 550
      & 24 & 21 & 32 & 32 & 22 & 148 & 10 & 25 & 32 & 101
    \end{tabular}

    \end{center}
    \caption{Number of unsolved tasks for the
  STRIPS domains from IPC1, IPC2, and IPC3.}
    \label{figure:experiments1-unsolved}
  \end{figure}

\includegraph[p]{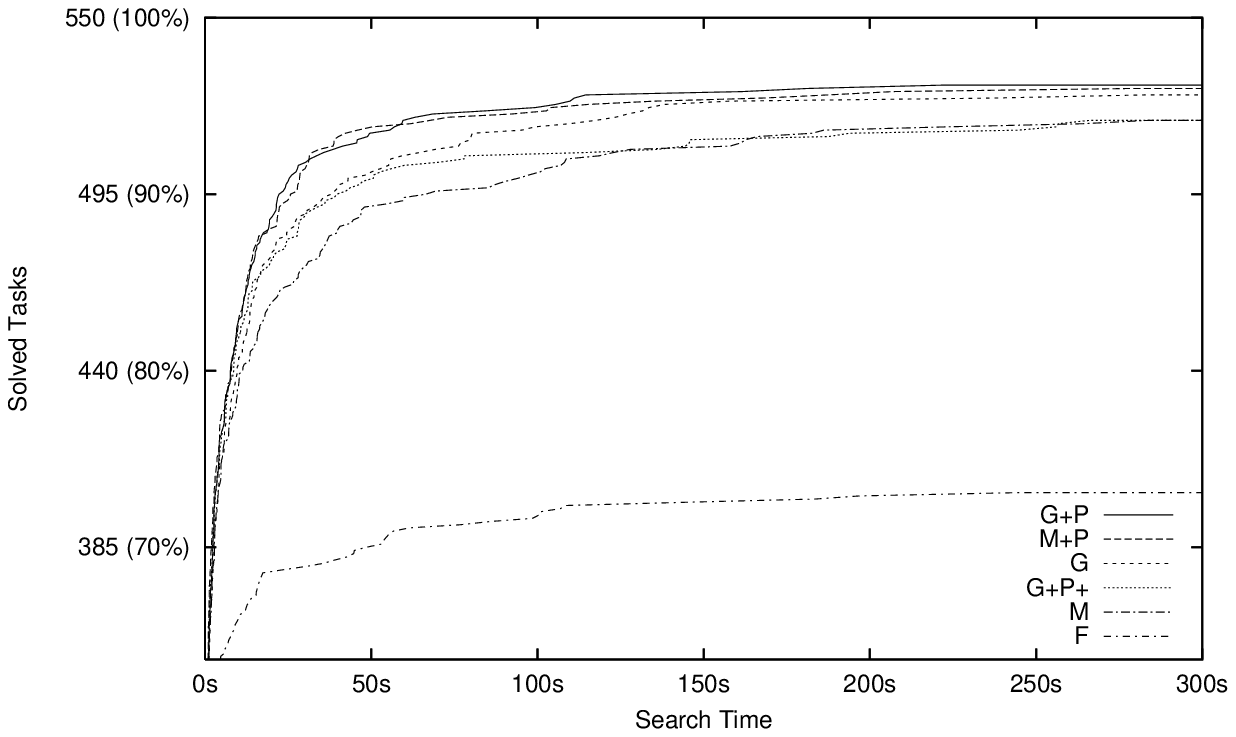}{Number of tasks solved
  vs.~runtime for the STRIPS domains from IPC1, IPC2 and IPC3. This
  graph shows the results for the various configurations of Fast
  Downward.}

\includegraph[p]{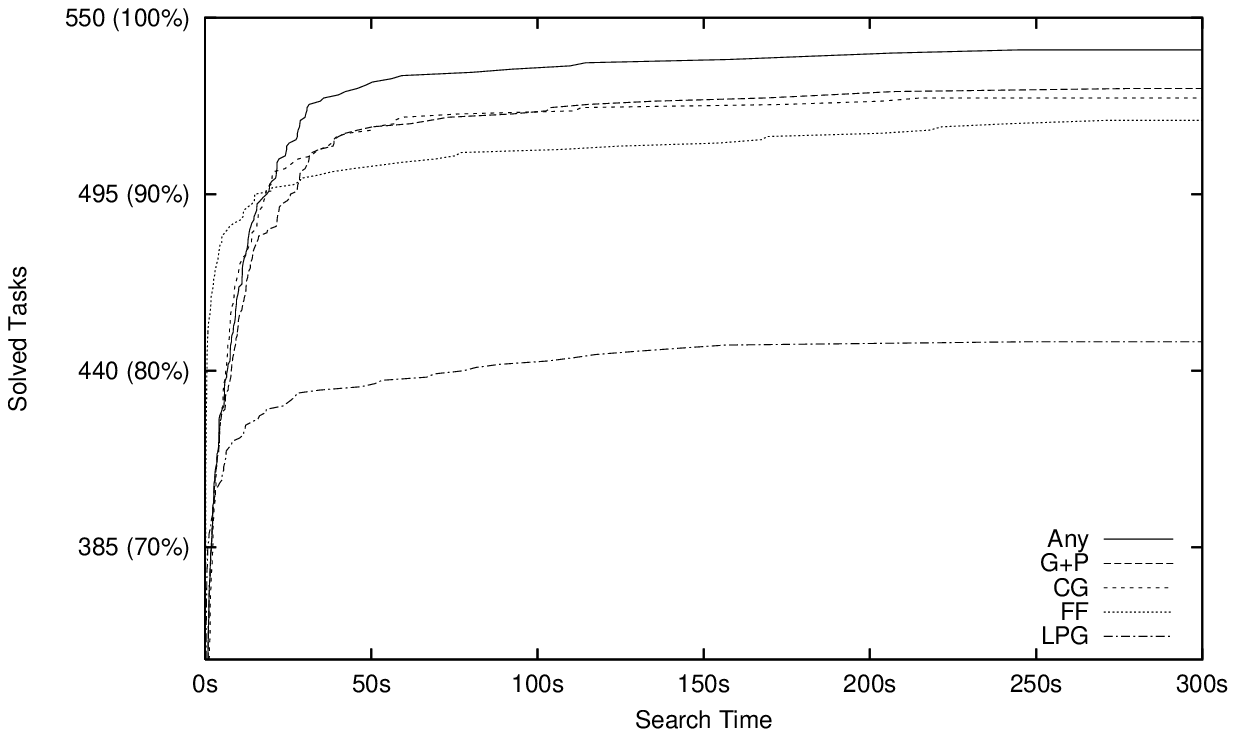}{Number of tasks solved
  vs.~runtime for the STRIPS domains from IPC1, IPC2 and IPC3. This
  graph shows the results for CG, FF and LPG and the hypothetical
  ``Any'' planner which always chooses the best configuration of Fast
  Downward. The result for greedy best-first search with helpful
  transitions is repeated for ease of comparison with
  Fig.~\ref{figure:experiment1a-graph}.}

Fig.~\ref{figure:experiments1-unsolved} shows the number of unsolved
tasks for each of the STRIPS domains from IPC1--3.
Figs.~\ref{figure:experiment1a-graph} and
\ref{figure:experiment1b-graph} show the number of tasks solved by
each planner within a given time bound between 0 and 300 seconds.
In addition to the six configurations of Fast Downward under
consideration, the table includes four other columns.

Under the heading ``Any'', we include results for a hypothetical
meta-planner that guesses the best of the six configuration of Fast
Downward for each input task and then executes Fast Downward with this
setting. Under the heading ``CG'', we report the results for our first
implementation of the causal graph heuristic
\cite{malte-icaps-cg}.\footnote{Apart from missing support for ADL and
  axioms, CG is very similar to Fast Downward using greedy best-first
  search and no preferred operators (configuration \textbf{G}). The
  translation and knowledge compilation components are essentially
  identical. The older search component mainly differs from Fast
  Downward in that it does not use deferred heuristic evaluation.}
Finally, ``FF'' and ``LPG'' refer to the well-known planners
\cite{ff,lpg} which won the fully-automated tracks of IPC2 and IPC3.
They were chosen for comparison on this benchmark set because they
showed the best performance by far of all publicly available planners
we experimented with. For LPG, which uses a randomized search
strategy, we attempted to solve each task five times and report the
median result.

The results show excellent performance of Fast Downward on this set of
benchmarks. Compared to CG, which was already shown to solve more
tasks than FF and LPG on this benchmark set \cite{malte-icaps-cg}, we
get another slight improvement for half of the planner configurations.
One of the configurations, multi-heuristic best-first search
using preferred operators, solves all benchmarks in all domains except
{\depot} and {\freecell}. Even more importantly, the number of tasks
not solved by any of the Fast Downward configurations is as small as 10.
Note that the planning competitions typically allowed a planner to
spend 30 minutes on each task; under these time constraints, we could
allocate five minutes to each of the six configurations of Fast
Downward, getting results which are at least as good as those reported
for the ``Any'' planner. Results might even be better under a cleverer
allocation scheme.

Even the configuration using focused iterative-broadening search
performs comparatively well on these benchmarks, although it cannot
compete with the other planners. Not surprisingly, this version of the
planner has difficulties in domains with many dead ends
(\freecell, \mystery, \mysteryprime) or where goal ordering is very
important (\blocksworld, \depot). It also fares comparatively badly in
domains with very large instances, namely {\logistics} (IPC1) and
\satellite. The reader should keep in mind that FF and LPG are
excellent planning systems; of all the other planners we experimented
with, including all those that were awarded prizes at the first three
planning competitions, none solved more benchmarks from this group
than focused iterative-broadening search.

The one domain that proves quite resistant to Fast Downward's solution
attempts in any configuration is {\depot}. As we already observed in
the initial experiments with the causal graph heuristic
\cite{malte-icaps-cg}, we believe that one key problem here is that
Fast Downward, unlike FF, does not use any goal ordering techniques,
which are very important in this domain. The fact that the domain
includes a \blocksworld-like subproblem is also problematic, as it
gives rise to very dense causal graphs as we demonstrated in Section
\ref{section:causal-graph-examples}.

\subsection{ADL Domains from IPC1--3}

Second, we present results for the ADL domains of the first three
planning competitions. This is a much smaller group than the previous,
including only four domains. This time, we cannot consider CG or LPG,
since neither CG nor the publicly available version of LPG supports
ADL domains. Therefore, we compare to FF exclusively. Again, we report
the number of unsolved tasks in each domain
(Fig.~\ref{figure:experiments2-unsolved}) and present graphs showing
how quickly the tasks are solved
(Figs.~\ref{figure:experiment2a-graph} and
\ref{figure:experiment2b-graph}).

  \begin{figure}[htbp]
    \begin{center}
    \newcommand{\heading}[1]{%
      \multicolumn{1}{c}{\makebox[0.6cm]{\textbf{#1}}}}
    \newcommand{\headingrule}[1]{%
      \multicolumn{1}{c|}{\makebox[0.6cm]{\textbf{#1}}}}
    \begin{tabular}{lr|cccccc|c|c}
      \heading{Domain}
    & \multicolumn{1}{c|}{\textbf{\#Tasks}}
    & \heading{G}
    & \heading{G+P}
    & \heading{G+P${}^+$}
    & \heading{M}
    & \heading{M+P}
    & \headingrule{F}
    & \headingrule{Any}
    & \heading{FF} \\[6pt]
    \assembly & 30 &
      28 & 27 & 25 & 3 & 0 & 30 & 0 & 0 \\
    \miconicsimple & 150 & 
      0 & 0 & 0 & 0 & 0 & 0 & 0 & 0 \\
    \miconicfull & 150 &
      9 & 8 & 9 & 9 & 8 & 90 & 6 & 12 \\
    \schedule & 150 &
      134 & 93 & 93 & 132 & 28 & 113 & 25 & 0 \\
    \textbf{Total} & 480
      & 171 & 128 & 127 & 144 & 36 & 233 & 31 & 12
    \end{tabular}

    \end{center}
    \caption{Number of unsolved tasks for the
  ADL domains from IPC1, IPC2 and IPC3.}
    \label{figure:experiments2-unsolved}
  \end{figure}

\includegraph[p]{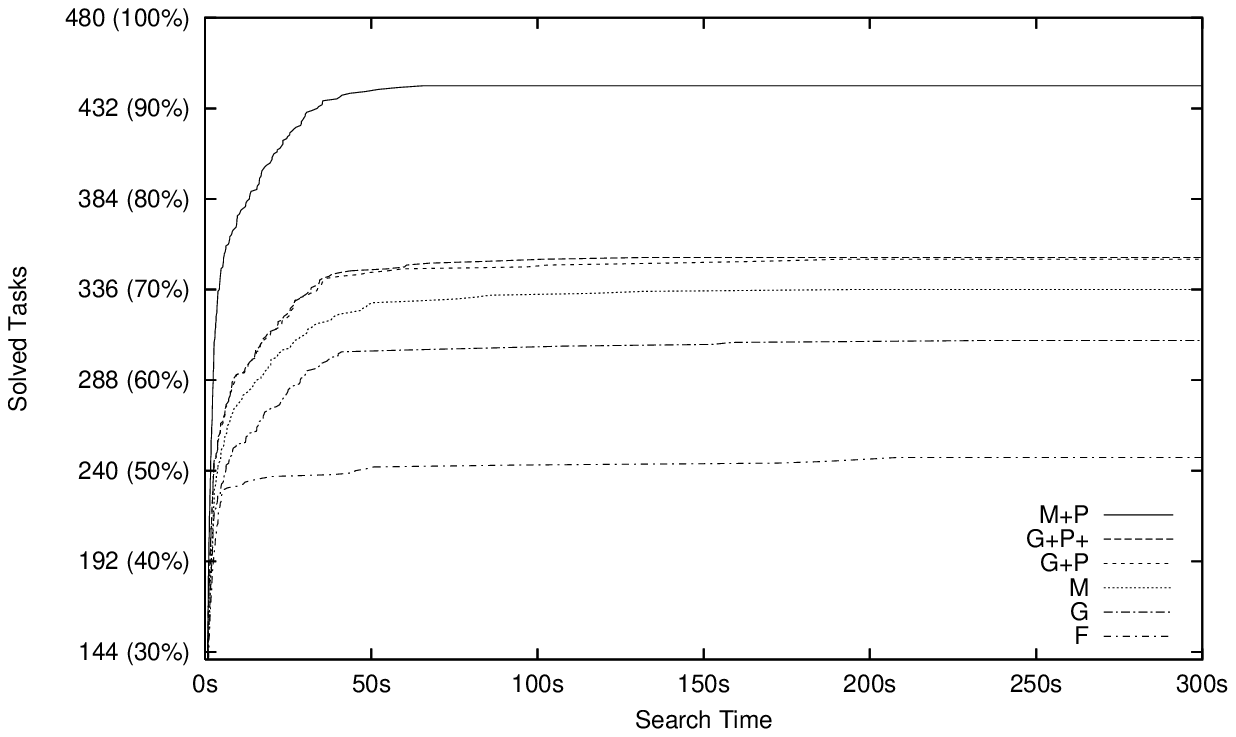}{Number of tasks solved
  vs.~runtime for the ADL domains from IPC1, IPC2 and IPC3. This
  graph shows the results for the various configurations of Fast
  Downward.}

\includegraph[p]{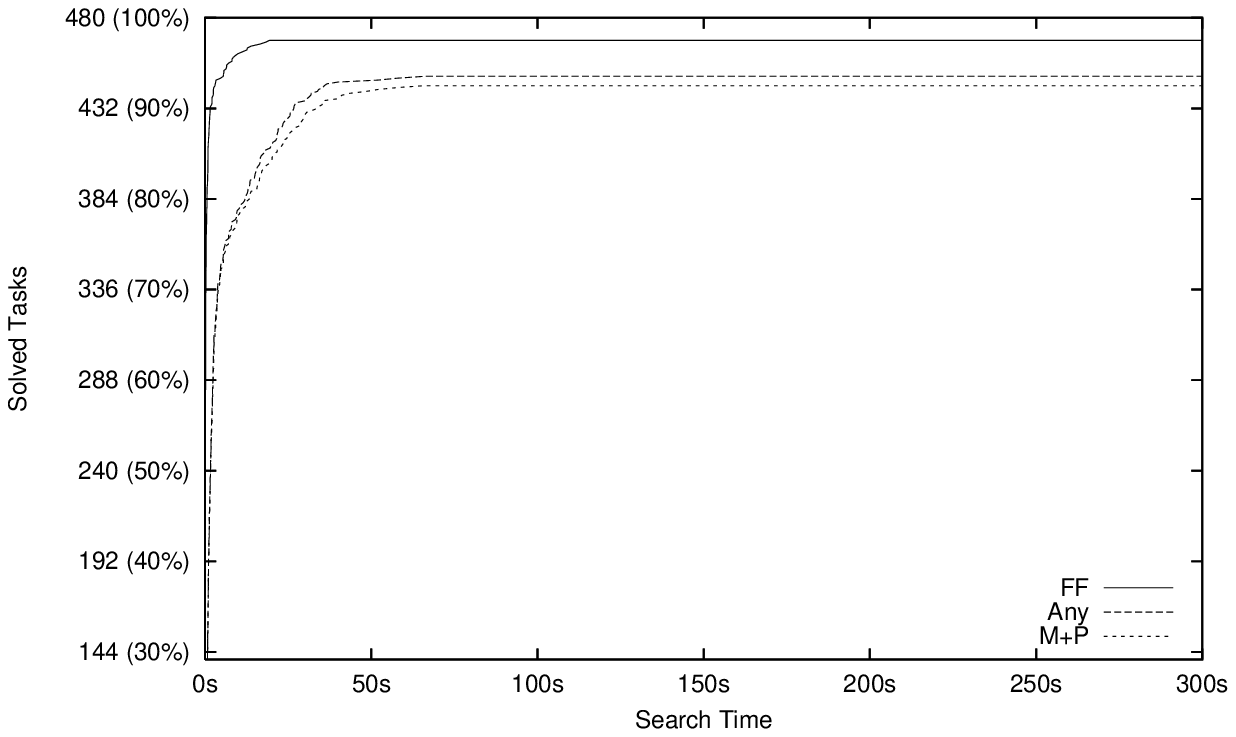}{Number of tasks solved
  vs.~runtime for the ADL domains from IPC1, IPC2 and IPC3. This
  graph shows the results for FF and the hypothetical
  ``Any'' planner which always chooses the best configuration of Fast
  Downward. The result for multi-heuristic best-first search with
  preferred operators is repeated for ease of comparison with
  Fig.~\ref{figure:experiment2a-graph}.}

These results do not look as good as for the first group of domains.
Results in both {\miconic} domains are good, even improving on those
of FF. However, greedy best-first search performs very badly in the
{\assembly} domain, and all configurations perform badly in the
{\schedule} domain. Currently, we have no good explanation for the
{\assembly} behaviour. For the {\schedule} domain, the weak performance
again seems to be related to missing goal ordering techniques: In many
{\schedule} tasks, several goals are defined for the same object which
can only be satisfied in a certain order. For instance, for objects
that should be cylindrical, polished and painted, these three goals
must be satisfied in precisely this order: making an object
cylindrical reverts the effects of polishing and painting, and
polishing reverts the effect of painting. Not recognising these
constraints, the heuristic search algorithm assumes to be close to the
goal when an object is already polished and painted but not
cylindrical, and is loathe to transform the object into cylindrical
shape because this would undo the already achieved goals. With some
rudimentary manual goal ordering, ignoring painting goals until all
other goals have been satisfied, the number of tasks not solved by
multi-heuristic best-first search with preferred operators drops from
28 to 3. These three failures appear to be due to the remaining ordering
problems with regard to cylindrical and polished objects.

\subsection{Domains from IPC4}

Third and finally, we present results for the IPC4 domains. Here, we
do not compare to FF: for these benchmarks, FF does not perform as
well as the best planners from the competition. Besides, several of
the IPC4 competitors are extensions of FF or hybrids using FF as part
of a bigger system, so FF-based planning is well-represented even if
we limit our attention to the IPC4 planners. For this comparison, we
chose the four most successful competition participants besides Fast
Downward, namely LPG-TD, SGPlan, Macro-FF and YAHSP \cite<cf.~the
results in>{ipc4}.
Similar to the previous two experiments, we report the number of
unsolved tasks in each domain
(Fig.~\ref{figure:experiments3-unsolved}) and present graphs showing
how quickly the tasks are solved
(Figs.~\ref{figure:experiment3a-graph} and
\ref{figure:experiment3b-graph}).

  \begin{figure}[htbp]
    \begin{center}
    \newcommand{\heading}[1]{%
      \multicolumn{1}{c}{\makebox[0.6cm]{\textbf{#1}}}}
    \newcommand{\headingrule}[1]{%
      \multicolumn{1}{c|}{\makebox[0.6cm]{\textbf{#1}}}}
    \begin{tabular}{lr|cccccc|c}
      \heading{Domain}
    & \multicolumn{1}{c|}{\textbf{\#Tasks}}
    & \heading{G}
    & \heading{G+P}
    & \heading{G+P${}^+$}
    & \heading{M}
    & \heading{M+P}
    & \headingrule{F}
    & \heading{Any} \\[6pt]
    \airport & 50
      & 28 & 30 & 17 & 18 & 14 &  0 &  0 \\
    \pipesworldnotankage & 50
      & 24 & 25 & 23 & 14 &  7 & 10 &  7 \\
    \pipesworldtankage & 50
      & 36 & 36 & 36 & 34 & 17 & 34 & 14 \\
    \opticaltelegraph & 48
      & 48 & 47 & 48 & 47 & 46 & 13 & 13 \\
    \philosophers & 48
      &  0 &  0 &  0 & 16 &  0 & 21 &  0 \\
    \psrsmall & 50
      &  0 &  0 &  0 &  0 &  0 &  1 &  0 \\
    \psrmiddle & 50
      &  0 &  0 &  0 &  0 &  0 & 22 &  0 \\
    \psrlarge & 50
      & 22 & 20 & 22 & 23 & 22 & 39 & 20 \\
    {\satellite} (IPC4) & 36
      &  8 &  0 &  0 &  8 &  3 & 22 &  0 \\
    \textbf{Total} & 432
      &166 &158 &146 &160 & 109&162 & 54
    \end{tabular}

    \bigskip

    \renewcommand{\heading}[1]{%
      \multicolumn{1}{c}{\textbf{#1}}}
    \renewcommand{\headingrule}[1]{%
      \multicolumn{1}{c|}{\textbf{#1}}}
    \begin{tabular}{l|cc|cccc}
      \headingrule{Domain}
    & \heading{FD}
    & \headingrule{FDD}
    & \heading{LPG-TD}
    & \heading{Macro-FF}
    & \heading{SGPlan}
    & \heading{YAHSP} \\[6pt]
    \airport
      &  0 &  0 &  7 & 30 &  6 & 17 \\
    \pipesworldnotankage
      & 11 &  7 & 10 & 12 &  0 &  0 \\
    \pipesworldtankage
      & 34 & 19 & 29 & 29 & 20 & 13 \\
    \opticaltelegraph
      & 22 & 22 & 37 & 31 & 29 & 36 \\
    \philosophers
      &  0 &  0 &  1 & 36 &  0 & 19 \\
    \psrsmall
      &  0 &  0 &  2 & 50 &  6 &  3 \\
    \psrmiddle
      &  0 &  0 &  0 & 19 &  4 & 50 \\
    \psrlarge
      & 22 & 22 & 50 & 50 & 39 & 50 \\
    {\satellite} (IPC4)
      &  0 &  3 &  1 &  0 &  6 &  0 \\
    \textbf{Total}
      & 89 & 73 &137 & 257&110 &188
    \end{tabular}

    \end{center}
    \caption{Number of unsolved tasks for the
  IPC4 domains. Results for the various configurations of Fast Downward
  are listed in the upper part, results for the competition
  participants in the lower part. ``FD'' and ``FDD'' denote the
  versions of Fast Downward that participated in IPC4 under the names
  ``Fast Downward'' and ``Fast Diagonally Downward''
  (cf.~Section~\ref{section:search}).}
    \label{figure:experiments3-unsolved}
  \end{figure}

\includegraph[p]{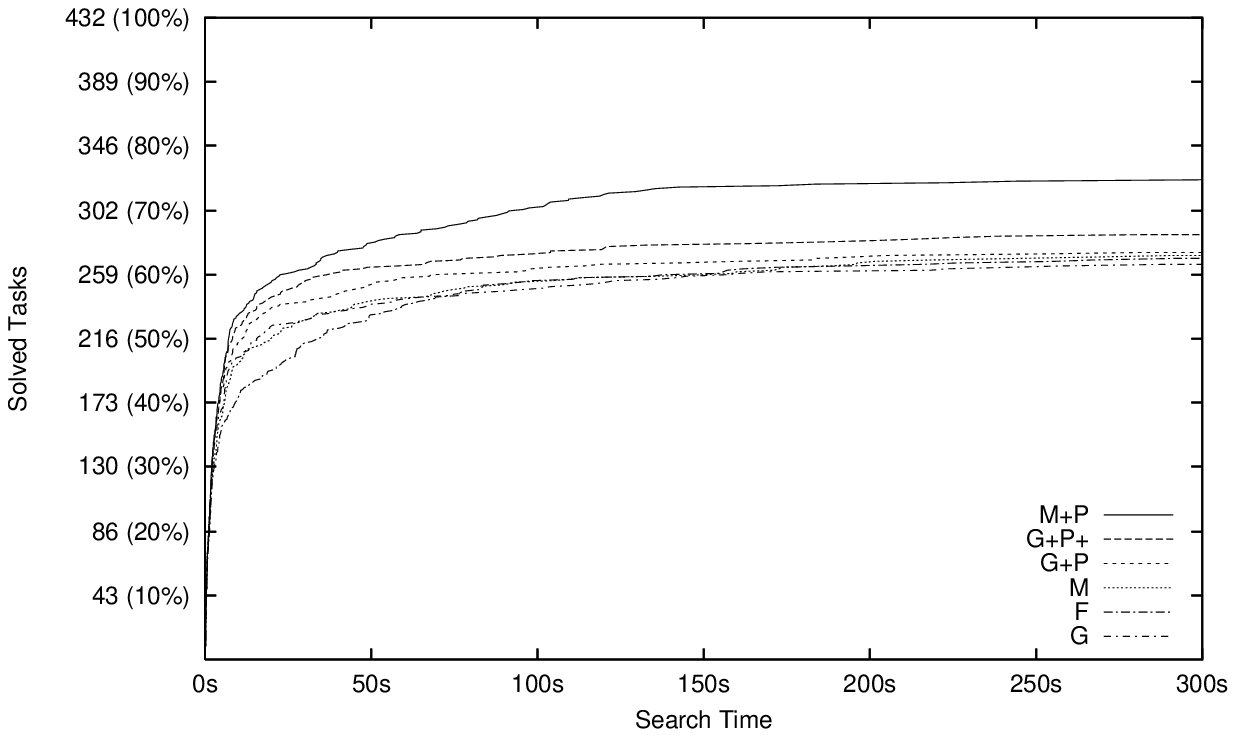}{Number of tasks solved
  vs.~runtime for the IPC4 domains. This graph shows the results for
  the various configurations of Fast Downward.}

\includegraph[p]{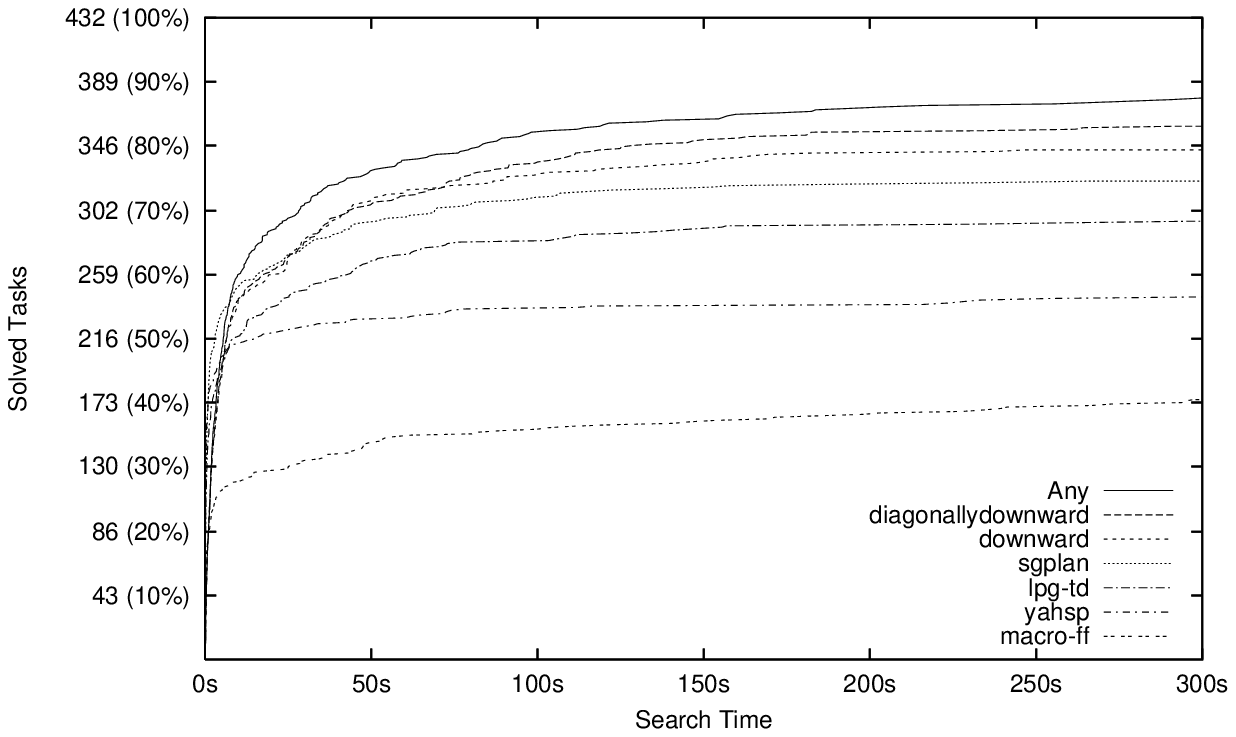}{Number of tasks solved
  vs.~runtime for the IPC4 domains. This graph shows the results for
  the hypothetical ``Any'' planner which always chooses the best
  configuration of Fast Downward, the competition configurations of
  Fast Downward and the best four other participants.}

Fast Downward is competitive with the other planners across domains,
and better than all others in some. The {\pipesworld} domains are the
only ones in which any of the other planners is noticeably better than the two
competition versions of Fast Downward. This is the case for YAHSP in
both {\pipesworld} domain variants and for SGPlan in {\pipesworldnotankage}.
The {\pipesworld} domain is not very hierarchical in nature; this
might be a domain where the decomposition approach of the causal graph
heuristic is not very appropriate. The results of the heuristic search
configurations in the {\opticaltelegraph} domain are extremely bad and
require further investigation.

Interestingly, focused iterative-broadening search performs very well
on some of the benchmarks from this suite. One of the reasons for this
is that in many of the tasks of the IPC4 suite, there are many
individual goals which are easy to serialize and can be
solved mostly independently.\footnote{We have devised an experiment
  which shows that if this property is artificially violated by a
  simple goal reformulation, the performance of the algorithm degrades
  quickly; see the electronic appendix for details.}  Comparing the
configuration \textbf{G} to \textbf{G + P${}^+$} and especially
\textbf{M} to \textbf{M + P}, we also observe that using preferred
operators is very useful for these benchmarks, even more so than in
the two previous experiments.

As a final remark, we observe that if we implemented the ``Any''
meta-planner by calling the six Fast Downward configurations in a
round-robin fashion, we would obtain a planning system that could
solve all but 54 of the IPC4 benchmarks within a $6 \cdot 5 = 30$
minute timeout. This is almost on par with the top performer of IPC4,
Fast Diagonally Downward, which solved all but 52 of the IPC4
benchmarks under the same timeout. Thus, this is a benchmark set for
which exploring different planner configurations definitely pays off.

\subsection{Conclusions from the Experiment}

How can we interpret these experimental results? Our first conclusion is
that Fast Downward is clearly competitive with the state of the art.
This is especially true for the configuration using multi-heuristic
best-first search with preferred operators (\textbf{M+P}), which
outperforms all competing planning systems both on the set of STRIPS
domains from IPC1--3 and on the domains from IPC4. If it were not for
the problems in the {\schedule} domain, the same would be true for the
remaining group of benchmarks, the ADL domains from IPC1--3.

With regard to the second objective of the investigation, evaluating
the relative strengths of the different planner configurations, the
\textbf{M+P} configuration emerges as a clear-cut winner. In 23 out of
29 domains, no other configuration solves more tasks, and unlike the
other configurations, there is only one domain (\opticaltelegraph) in
which it performs very badly. We conclude that both multi-heuristic
best-first search and the use of preferred operators are promising
extensions to heuristic planners.

This is particularly true for preferred operators. Indeed, after the
\textbf{M+P} configuration, the two variants of greedy best-first
search with preferred operators show the next best overall
performance, both in terms of the number of domains where they are among
the top performers and in terms of the total number of tasks solved.
Comparing \textbf{G} to \textbf{G+P}, there are ten domains in which
the variant using preferred operators solves more tasks than the one
not using them; the opposite is true in five domains. Comparing
\textbf{M} to \textbf{M+P}, the difference is even more striking, with
the preferred operator variant outperforming the other in fifteen
domains, while being worse in two (in both of which it only solves one
task less).  These are convincing arguments for the use of preferred
operators.

\section{Summary and Discussion}
\label{section:discussion}

Before we turn to discussion, let us briefly summarize the
contributions of this article. As a motivating starting point, we
explained that planning tasks often exhibit a simpler structure if
expressed with \emph{multi-valued state variables}, rather than the
traditional propositional representations. We then introduced
\emph{Fast Downward}, a planning system based on the idea of
converting tasks into a multi-valued formalism and exploiting the
causal information underlying such encodings.

Fast Downward processes PDDL planning tasks in three stages. We
skipped the first of these stages, \emph{translation}, which
automatically transforms a PDDL task into an equivalent multi-valued
planning task with a nicer causal structure. We explained the inner
workings of the second stage, \emph{knowledge compilation},
demonstrating in depth what kind of knowledge the planner extracts
from the problem representation, discussing \emph{causal graphs},
\emph{domain transition graphs}, \emph{successor generators} and
\emph{axiom evaluators}. During our discussion of Fast Downward's
\emph{search} component, we introduced its heuristic search
algorithms, which use the technique of \emph{deferred heuristic
  evaluation} to reduce the number of states for which a heuristic
goal distance estimate must be computed. In addition to greedy
best-first search, Fast Downward employs the \emph{multi-heuristic
best-first search} algorithm to usefully integrate the information
of two heuristic estimators, namely the \emph{causal graph heuristic} and
\emph{FF heuristic}. Both heuristic search algorithms can utilize
preference information about operators. We also introduced Fast
Downward's experimental \emph{focused iterative-broadening search}
algorithm, which is based on the idea of pruning the set of operators
to only consider those successor states which are likely to lead
towards a specific goal.

We thus tried to give a complete account of the Fast Downward
planning system's approach to solving multi-valued planning tasks,
including its motivation, architecture, and algorithmic
foundations. In the previous section, we demonstrated its empirical
behaviour, showing good performance across the whole range of
propositional benchmarks from the previous planning competitions.

Among all the novel algorithms and search enhancements discussed in
this article, there are two aspects of Fast Downward which we consider
of central importance and which we would like to emphasize.
One of them is the use of multi-valued state variables for PDDL-style
planning. We believe that multi-valued representations are much more
structured and hence much more amenable to automated reasoning --- be
it for the purposes of heuristic evaluation, problem decomposition, or
other aspects of planning such as goal ordering or extraction of
landmarks.
The other central idea is the use of hierarchical decompositions
within a heuristic planning framework. Hierarchical approaches to
domain-independent planning have a considerable potential, but since
the work of Knoblock \citeyear{knoblock-abstractions} and Bacchus and
Yang \citeyear{bacchus-downward-refinement}, little work has been
published. With Fast Downward, we hope to renew interest in this area,
which we believe to be a very promising ground for further advances in
automated planning.

For the future, there are several aspects of Fast Downward that we
would like to investigate further. First, we intend to experiment with
other search techniques along the lines of focused
iterative-broadening search, which emphasize heuristically evaluating
operator usefulness rather than heuristically evaluating states.

Second, we would like to come up with an efficient heuristic for
multi-valued planning tasks which does not require pruning cycles of
the causal graph. Initial experiments in this direction have shown
that it is difficult to achieve this goal without losing the
performance of Fast Downward's heuristic estimator, but perhaps better
heuristic accuracy can outweigh worse per-state performance in many
cases.

Third, we want to investigate in how far the performance of the
planner could be improved by encoding some domains differently. In
some cases, merging a set of state variables which are closely
interrelated into a single state variable whose domain is the product
of the domains of the original state variables might be beneficial.
Also, we want to test if hand-tailored encodings lead to better
performance than automatically derived ones, and if so, how large the
performance gap is.

Fourth and finally, we would like to evaluate the
behaviour of the causal graph heuristic in specific planning domains
both empirically and theoretically, following Hoffmann's work on the
FF heuristic \cite{joerg-topology,hoffmann:aips-02,hoffmann:jair2005}.
Hopefully, this will give some indication when we can expect good
performance from the causal graph heuristic and when it is advisable
to look for other approaches.

\section*{Acknowledgements}
The author wishes to thank Silvia Richter, the other member of the Fast
Downward team at the 4th International Planning Competition, for her
part in implementing the planner and for valuable advice before,
throughout, and after the competition. She also deserves thanks for
helping out with the experiments, for proof-reading this article,
and for suggesting a number of improvements.

The anonymous reviewers of the article and the handling editor, Maria
Fox, made a number of useful suggestions that led to significant
improvements.

This work was partly supported by the German Research Council (DFG)
within the Graduate Programme ``Mathematical Logic and Applications''
and as part of the Transregional Collaborative Research Centre
``Automatic Verification and Analysis of Complex Systems'' (SFB/TR 14
AVACS). See \texttt{www.avacs.org} for more information.

\bibliography{abbrv,literatur,crossref}

\begin{thebibliography}{}

\bibitem[\protect\BCAY{Bacchus\ \BBA\ Yang}{Bacchus\ \BBA\
  Yang}{1994}]{bacchus-downward-refinement}
Bacchus, F.\BBACOMMA\  \BBA\ Yang, Q. \BBOP1994\BBCP.
\newblock \BBOQ Downward refinement and the efficiency of hierarchical problem
  solving\BBCQ\
\newblock {\Bem Artificial Intelligence}, {\Bem 71\/}(1), 43--100.

\bibitem[\protect\BCAY{B{\"a}ckstr{\"o}m\ \BBA\ Nebel}{B{\"a}ckstr{\"o}m\ \BBA\
  Nebel}{1995}]{complexity-sas}
B{\"a}ckstr{\"o}m, C.\BBACOMMA\  \BBA\ Nebel, B. \BBOP1995\BBCP.
\newblock \BBOQ Complexity results for {SAS$^{+}$} planning\BBCQ\
\newblock {\Bem Computational Intelligence}, {\Bem 11\/}(4), 625--655.

\bibitem[\protect\BCAY{Bonet\ \BBA\ Geffner}{Bonet\ \BBA\ Geffner}{2001}]{hsp}
Bonet, B.\BBACOMMA\  \BBA\ Geffner, H. \BBOP2001\BBCP.
\newblock \BBOQ Planning as heuristic search\BBCQ\
\newblock {\Bem Artificial Intelligence}, {\Bem 129\/}(1), 5--33.

\bibitem[\protect\BCAY{Brafman\ \BBA\ Domshlak}{Brafman\ \BBA\
  Domshlak}{2003}]{brafman-domshlak:jair2003}
Brafman, R.~I.\BBACOMMA\  \BBA\ Domshlak, C. \BBOP2003\BBCP.
\newblock \BBOQ Structure and complexity in planning with unary operators\BBCQ\
\newblock {\Bem Journal of Artificial Intelligence Research}, {\Bem 18},
  315--349.

\bibitem[\protect\BCAY{Bylander}{Bylander}{1994}]{complexity-bylander}
Bylander, T. \BBOP1994\BBCP.
\newblock \BBOQ The computational complexity of propositional {STRIPS}
  planning\BBCQ\
\newblock {\Bem Artificial Intelligence}, {\Bem 69\/}(1--2), 165--204.

\bibitem[\protect\BCAY{Domshlak\ \BBA\ Brafman}{Domshlak\ \BBA\
  Brafman}{2002}]{carmel-unary}
Domshlak, C.\BBACOMMA\  \BBA\ Brafman, R.~I. \BBOP2002\BBCP.
\newblock \BBOQ Structure and complexity in planning with unary operators\BBCQ\
\newblock In Ghallab, M., Hertzberg, J., \BBA\ Traverso, P.\BEDS, {\Bem
  Proceedings of the Sixth International Conference on Artificial Intelligence
  Planning and Scheduling ({AIPS} 2002)}, \BPGS\ 34--43. AAAI Press.

\bibitem[\protect\BCAY{Domshlak\ \BBA\ Dinitz}{Domshlak\ \BBA\
  Dinitz}{2001}]{carmel-ecp99}
Domshlak, C.\BBACOMMA\  \BBA\ Dinitz, Y. \BBOP2001\BBCP.
\newblock \BBOQ Multi-agent off-line coordination: {S}tructure and
  complexity\BBCQ\
\newblock In Cesta, A.\BBACOMMA\  \BBA\ Borrajo, D.\BEDS, {\Bem Pre-proceedings
  of the Sixth European Conference on Planning ({ECP}'01)}, \BPGS\ 277--288,
  Toledo, Spain.

\bibitem[\protect\BCAY{Dowling\ \BBA\ Gallier}{Dowling\ \BBA\
  Gallier}{1984}]{horn-marking}
Dowling, W.~F.\BBACOMMA\  \BBA\ Gallier, J.~H. \BBOP1984\BBCP.
\newblock \BBOQ Linear-time algorithms for testing the satisfiability of
  propositional {Horn} formulae\BBCQ\
\newblock {\Bem Journal of Logic Programming}, {\Bem 1\/}(3), 367--383.

\bibitem[\protect\BCAY{Edelkamp\ \BBA\ Helmert}{Edelkamp\ \BBA\
  Helmert}{1999}]{minimize}
Edelkamp, S.\BBACOMMA\  \BBA\ Helmert, M. \BBOP1999\BBCP.
\newblock \BBOQ Exhibiting knowledge in planning problems to minimize state
  encoding length\BBCQ\
\newblock In Fox, M.\BBACOMMA\  \BBA\ Biundo, S.\BEDS, {\Bem Recent Advances in
  {AI} Planning. 5th European Conference on Planning ({ECP}'99)},
  \lowercase{\BVOL}\ 1809 of {\Bem Lecture Notes in Artificial Intelligence},
  \BPGS\ 135--147, New York. Springer-Verlag.

\bibitem[\protect\BCAY{Edelkamp\ \BBA\ Hoffmann}{Edelkamp\ \BBA\
  Hoffmann}{2004}]{pddl2.2}
Edelkamp, S.\BBACOMMA\  \BBA\ Hoffmann, J. \BBOP2004\BBCP.
\newblock \BBOQ {PDDL2.2}: {The} language for the classical part of the 4th
  {International} {Planning} {Competition}\BBCQ\
\newblock \BTR\ 195, Albert-Ludwigs-Universit{\"a}t Freiburg, Institut f{\"u}r
  Informatik.

\bibitem[\protect\BCAY{Fox\ \BBA\ Long}{Fox\ \BBA\ Long}{2003}]{pddl2}
Fox, M.\BBACOMMA\  \BBA\ Long, D. \BBOP2003\BBCP.
\newblock \BBOQ {PDDL2.1}: {An} extension to {PDDL} for expressing temporal
  planning domains\BBCQ\
\newblock {\Bem Journal of Artificial Intelligence Research}, {\Bem 20},
  61--124.

\bibitem[\protect\BCAY{Garey\ \BBA\ Johnson}{Garey\ \BBA\
  Johnson}{1979}]{garey-johnson}
Garey, M.~R.\BBACOMMA\  \BBA\ Johnson, D.~S. \BBOP1979\BBCP.
\newblock {\Bem Computers and Intractability --- {A} Guide to the Theory of
  NP-Completeness}.
\newblock Freeman.

\bibitem[\protect\BCAY{Gerevini, Saetti,\ \BBA\ Serina}{Gerevini
  et~al.}{2003}]{lpg}
Gerevini, A., Saetti, A., \BBA\ Serina, I. \BBOP2003\BBCP.
\newblock \BBOQ Planning through stochastic local search and temporal action
  graphs in {LPG}\BBCQ\
\newblock {\Bem Journal of Artificial Intelligence Research}, {\Bem 20},
  239--290.

\bibitem[\protect\BCAY{Ginsberg\ \BBA\ Harvey}{Ginsberg\ \BBA\
  Harvey}{1992}]{ginsberg-harvey:aij92}
Ginsberg, M.~L.\BBACOMMA\  \BBA\ Harvey, W.~D. \BBOP1992\BBCP.
\newblock \BBOQ Iterative broadening\BBCQ\
\newblock {\Bem Artificial Intelligence}, {\Bem 55}, 367--383.

\bibitem[\protect\BCAY{Helmert}{Helmert}{2004}]{malte-icaps-cg}
Helmert, M. \BBOP2004\BBCP.
\newblock \BBOQ A planning heuristic based on causal graph analysis\BBCQ\
\newblock In Zilberstein, S., Koehler, J., \BBA\ Koenig, S.\BEDS, {\Bem
  Proceedings of the Fourteenth International Conference on Automated Planning
  and Scheduling ({ICAPS} 2004)}, \BPGS\ 161--170. AAAI Press.

\bibitem[\protect\BCAY{Hoffmann}{Hoffmann}{2001}]{joerg-topology}
Hoffmann, J. \BBOP2001\BBCP.
\newblock \BBOQ Local search topology in planning benchmarks: {A}n empirical
  analysis\BBCQ\
\newblock In Nebel, B.\BED, {\Bem Proceedings of the 17th International Joint
  Conference on Artificial Intelligence ({IJCAI}'01)}, \BPGS\ 453--458. Morgan
  Kaufmann.

\bibitem[\protect\BCAY{Hoffmann}{Hoffmann}{2002}]{hoffmann:aips-02}
Hoffmann, J. \BBOP2002\BBCP.
\newblock \BBOQ Local search topology in planning benchmarks: {A} theoretical
  analysis\BBCQ\
\newblock In Ghallab, M., Hertzberg, J., \BBA\ Traverso, P.\BEDS, {\Bem
  Proceedings of the Sixth International Conference on Artificial Intelligence
  Planning and Scheduling ({AIPS} 2002)}, \BPGS\ 92--100. AAAI Press.

\bibitem[\protect\BCAY{Hoffmann}{Hoffmann}{2005}]{hoffmann:jair2005}
Hoffmann, J. \BBOP2005\BBCP.
\newblock \BBOQ Where `ignoring delete lists' works: {Local} search topology in
  planning benchmarks\BBCQ\
\newblock {\Bem Journal of Artificial Intelligence Research}, {\Bem 24},
  685--758.

\bibitem[\protect\BCAY{Hoffmann\ \BBA\ Edelkamp}{Hoffmann\ \BBA\
  Edelkamp}{2005}]{ipc4}
Hoffmann, J.\BBACOMMA\  \BBA\ Edelkamp, S. \BBOP2005\BBCP.
\newblock \BBOQ The deterministic part of {IPC-4}: {A}n overview\BBCQ\
\newblock {\Bem Journal of Artificial Intelligence Research}, {\Bem 24},
  519--579.

\bibitem[\protect\BCAY{Hoffmann\ \BBA\ Nebel}{Hoffmann\ \BBA\ Nebel}{2001}]{ff}
Hoffmann, J.\BBACOMMA\  \BBA\ Nebel, B. \BBOP2001\BBCP.
\newblock \BBOQ The {FF} planning system: {Fast} plan generation through
  heuristic search\BBCQ\
\newblock {\Bem Journal of Artificial Intelligence Research}, {\Bem 14},
  253--302.

\bibitem[\protect\BCAY{Jonsson\ \BBA\ B{\"a}ckstr{\"o}m}{Jonsson\ \BBA\
  B{\"a}ckstr{\"o}m}{1995}]{jonsson-backstrom:ewsp95}
Jonsson, P.\BBACOMMA\  \BBA\ B{\"a}ckstr{\"o}m, C. \BBOP1995\BBCP.
\newblock \BBOQ Incremental planning\BBCQ\
\newblock In Ghallab, M.\BBACOMMA\  \BBA\ Milani, A.\BEDS, {\Bem New Directions
  in {AI} Planning: {EWSP} '95 --- 3rd {European} Workshop on Planning},
  \lowercase{\BVOL}~31 of {\Bem Frontiers in Artificial Intelligence and
  Applications}, \BPGS\ 79--90, Amsterdam. IOS Press.

\bibitem[\protect\BCAY{Jonsson\ \BBA\ B{\"a}ckstr{\"o}m}{Jonsson\ \BBA\
  B{\"a}ckstr{\"o}m}{1998a}]{jonsson-sas}
Jonsson, P.\BBACOMMA\  \BBA\ B{\"a}ckstr{\"o}m, C. \BBOP1998a\BBCP.
\newblock \BBOQ State-variable planning under structural restrictions:
  {Algorithms} and complexity\BBCQ\
\newblock {\Bem Artificial Intelligence}, {\Bem 100\/}(1--2), 125--176.

\bibitem[\protect\BCAY{Jonsson\ \BBA\ B{\"a}ckstr{\"o}m}{Jonsson\ \BBA\
  B{\"a}ckstr{\"o}m}{1998b}]{jonsson-3s}
Jonsson, P.\BBACOMMA\  \BBA\ B{\"a}ckstr{\"o}m, C. \BBOP1998b\BBCP.
\newblock \BBOQ Tractable plan existence does not imply tractable plan
  generation\BBCQ\
\newblock {\Bem Annals of Mathematics and Artificial Intelligence}, {\Bem
  22\/}(3), 281--296.

\bibitem[\protect\BCAY{Joslin\ \BBA\ Roach}{Joslin\ \BBA\
  Roach}{1989}]{joslin-roach:aij89}
Joslin, D.\BBACOMMA\  \BBA\ Roach, J. \BBOP1989\BBCP.
\newblock \BBOQ A theoretical analysis of conjunctive-goal problems\BBCQ\
\newblock {\Bem Artificial Intelligence}, {\Bem 41\/}(1), 97--106.
\newblock Research Note.

\bibitem[\protect\BCAY{Knoblock}{Knoblock}{1994}]{knoblock-abstractions}
Knoblock, C.~A. \BBOP1994\BBCP.
\newblock \BBOQ Automatically generating abstractions for planning\BBCQ\
\newblock {\Bem Artificial Intelligence}, {\Bem 68\/}(2), 243--302.

\bibitem[\protect\BCAY{Korf}{Korf}{1987}]{korf:aij-1987}
Korf, R.~E. \BBOP1987\BBCP.
\newblock \BBOQ Planning as search: {A} quantitative approach\BBCQ\
\newblock {\Bem Artificial Intelligence}, {\Bem 33\/}(1), 65--88.

\bibitem[\protect\BCAY{Lowerre}{Lowerre}{1976}]{beam-search}
Lowerre, B.~T. \BBOP1976\BBCP.
\newblock {\Bem The \textsc{HARPY} Speech Recognition System}.
\newblock Ph.D.\ thesis, Computer Science Department, Carnegie-Mellon
  University, Pittsburgh, Pennsylvania.

\bibitem[\protect\BCAY{Newell\ \BBA\ Simon}{Newell\ \BBA\
  Simon}{1963}]{newell-simon:1963}
Newell, A.\BBACOMMA\  \BBA\ Simon, H.~A. \BBOP1963\BBCP.
\newblock \BBOQ {GPS}: {A} program that simulates human thought\BBCQ\
\newblock In Feigenbaum, E.~A.\BBACOMMA\  \BBA\ Feldman, J.\BEDS, {\Bem
  Computers and Thought}, \BPGS\ 279--293. Oldenbourg.

\bibitem[\protect\BCAY{Russell\ \BBA\ Norvig}{Russell\ \BBA\
  Norvig}{2003}]{russell-norvig}
Russell, S.\BBACOMMA\  \BBA\ Norvig, P. \BBOP2003\BBCP.
\newblock {\Bem Artificial {I}ntelligence --- {A} Modern Approach}.
\newblock Prentice Hall.

\bibitem[\protect\BCAY{Sacerdoti}{Sacerdoti}{1974}]{sacerdoti:aij74}
Sacerdoti, E.~D. \BBOP1974\BBCP.
\newblock \BBOQ Planning in a hierarchy of abstraction spaces\BBCQ\
\newblock {\Bem Artificial Intelligence}, {\Bem 5}, 115--135.

\bibitem[\protect\BCAY{Tenenberg}{Tenenberg}{1991}]{tenenberg:1991}
Tenenberg, J.~D. \BBOP1991\BBCP.
\newblock \BBOQ Abstraction in planning\BBCQ\
\newblock In Allen, J.~F., Kautz, H.~A., Pelavin, R.~N., \BBA\ Tenenberg,
  J.~D., {\Bem Reasoning About Plans}, \BCH~4, \BPGS\ 213--283. Morgan
  Kaufmann, San Mateo.

\bibitem[\protect\BCAY{{van den Briel}, Vossen,\ \BBA\ Kambhampati}{{van den
  Briel} et~al.}{2005}]{menkes-ilp}
{van den Briel}, M., Vossen, T., \BBA\ Kambhampati, S. \BBOP2005\BBCP.
\newblock \BBOQ Reviving integer programming approaches for {AI} planning: {A}
  branch-and-cut framework\BBCQ\
\newblock In Biundo, S., Myers, K., \BBA\ Rajan, K.\BEDS, {\Bem Proceedings of
  the Fifteenth International Conference on Automated Planning and Scheduling
  ({ICAPS} 2005)}, \BPGS\ 310--319. AAAI Press.

\bibitem[\protect\BCAY{Williams\ \BBA\ Nayak}{Williams\ \BBA\
  Nayak}{1997}]{williams-nayak}
Williams, B.~C.\BBACOMMA\  \BBA\ Nayak, P.~P. \BBOP1997\BBCP.
\newblock \BBOQ A reactive planner for a model-based executive\BBCQ\
\newblock In Pollack, M.~E.\BED, {\Bem Proceedings of the 15th International
  Joint Conference on Artificial Intelligence ({IJCAI}'97)}, \BPGS\ 1178--1195.
  Morgan Kaufmann.

\bibitem[\protect\BCAY{Yoshizumi, Miura,\ \BBA\ Ishida}{Yoshizumi
  et~al.}{2000}]{astar-partial-expansion}
Yoshizumi, T., Miura, T., \BBA\ Ishida, T. \BBOP2000\BBCP.
\newblock \BBOQ A${}^*$ with partial expansion for large branching factor
  problems\BBCQ\
\newblock In Kautz, H.\BBACOMMA\  \BBA\ Porter, B.\BEDS, {\Bem Proceedings of
  the Seventeenth National Conference on Artificial Intelligence
  ({AAAI}-2000)}, \BPGS\ 923--929. AAAI Press.

\end{thebibliography}
\bibliographystyle{theapa}

\end{document}